\documentclass[12pt]{article}

\usepackage{graphicx}
\usepackage{hyperref}
\usepackage{natbib}
\usepackage{multirow}
\usepackage{amsmath}
\usepackage{booktabs}
\usepackage{tabularray}
\usepackage{makecell}
\usepackage[table]{xcolor}
\usepackage{subcaption}
\usepackage[a4paper, total={6.3in, 8in}]{geometry}
\usepackage{times} 
\usepackage{amssymb}

\begin{document}

\title{Type Diversity Enables Transformers to \\ Generalise Compositionally}

\author{Anssi Moisio$^{1}$, Mathias Creutz$^{2}$, Mikko Kurimo$^{1}$ \\
\small$^{1}$Aalto University, $^{2}$University of Helsinki}

\date{}

\maketitle

\begin{abstract}
Compositional generalisation has been divided into lexical and structural generalisation.
Previous work has found that structural generalisation is harder than lexical for Transformers. We propose that this difference is not inherent to Transformers, but due to the high diversity of lexical types and low diversity of structural types in the specific datasets of these previous works. By \emph{type diversity} we mean the number of different constructors of that type, instead of, for example, the specific word combinations that might populate the structure.
To test this, we vary the amounts of type diversity of lexical and structural types in previously published datasets. We create linguistically diverse variants of the COGS and SLOG datasets using Grammatical Framework. 
We find that type diversity correlates with compositional generalisation equally in lexical and structural test cases,
supporting our hypothesis.
We note a contradiction with the proposition in previous work that \emph{compound divergence} explains the difficulty in compositional generalisation tasks. We further investigate the effects of other dataset properties on compositional generalisation, such as the diversity of types other than the novel test structure, and surface properties of the logical semantics format.

\end{abstract}

\section{Introduction}

The largest Transformer-based \citep{vaswani2017attention} language models
are increasingly competent in a wide variety of tasks, including generalising to tasks and languages that are not in their training set. On the other hand, multiple studies have shown that small Transformer models do not generalise to new linguistic structures accurately when trained on small diagnostic datasets \citep{kim-linzen-2020-cogs,yao-koller-2022-structural,li-etal-2023-slog,jabbar-etal-2025-distinguishing}.
Perhaps the most obvious explanation for this discrepancy would be the difference in \emph{scale}, which conventionally refers to the size of the training dataset, the amount of compute, and the number of learnable parameters.

However, these three factors are not the full picture of scale, and recent studies have pointed out \emph{data quality} as a fourth aspect \citep{chen-etal-2025-revisiting}. We take data quality, specifically linguistic diversity, as our focus in this study. Diversity typically comes along when the dataset size is scaled up, but it can also be scaled up independent of dataset size. Our main hypothesis is that the largest factor that explains the poor structural generalisation results is not an ineptitude of the Transformer itself nor the tripartite scale but instead a specific kind linguistic diversity, which we call \emph{type diversity}, of the training dataset. 
We define type diversity of a linguistic type, e.g. noun phrase, as the number of different constructor functions that build trees of that type. For example, the noun phrases `a cat' and `the mice' are built by one constructor, which builds simple NPs from common nouns and determiners, and `a cat on a mat' and `mice in the attic' are built by another constructor, which builds NPs from the simple NPs and prepositional phrases.

We test our hypothesis by adding natural linguistic diversity, such as adjectives, plural forms of nouns, and verb tenses into the popular COGS  dataset \citep{kim-linzen-2020-cogs} and the SLOG variant \citep{li-etal-2023-slog} of it. Besides these quasi-natural datasets, we test our hypothesis in the purely artificial SCAN dataset \citep{lake2018generalization} by generating over a hundred variants of it with differing amounts of type diversity. 

While assessing the impact of type diversity, we also report findings on other data factors.
Previous work has shown that incidental surface details can also have a surprisingly large impact on the results in a generalisation benchmark \citep{wu-etal-2023-recogs}.
We map the effect of certain concrete and more incidental properties, such as the order of conjuncts in the logical semantics. 

One previous method to assess the  properties of compositional generalisation (CG) datasets is called distribution-based compositionality assessment, or DBCA \citep{keysers2019measuring}. This method aims to evaluate `compositionality' based on the distributions of atomic units on one hand, and combinations of atoms on the other hand.
The recently found evidence that diversity helps CG, including our findings on type diversity, contradicts the observation that compound divergence correlates negatively with Transformer performance.

The paper is structured as follows. 
First are brief introductions to Grammatical Framework (Section~\ref{sec:gf}), which is used both as a theoretical framework to define type diversity and as a practical tool to generate datasets, and  to CG (Section~\ref{sec:compositionality}).
Section~\ref{sec:why-is-structural-hard} reviews previous work that has shown that lexical generalisation is easy while structural is hard for Transformers.
%
Section~\ref{sec:hypothesis} defines type diversity and reviews relevant previous work.
Section~\ref{sec:data} describes the data generation methods.
Section~\ref{sec:expts-cogs}
shows, using diversified variants of the COGS dataset, that adding new constructors for noun phrases
helps in a structural generalisation task.
Subsection~\ref{sec:expts-scan} describes experiments in the simpler SCAN domain, which more precisely map the relationship between
lexical/structural generalisation and type diversity. Subsection~\ref{sec:expts-dbca} reports the relationship of type diversity and the DBCA metric of \citet{keysers2019measuring}.
Using diversified variants of the SLOG dataset, Section~\ref{sec:expts-slog} outlines how some of the concrete syntax properties, such as the order of semantic terms, affect the results too. Finally, we discuss the pitfalls of assessing CG with diagnostic datasets in Section~\ref{sec:discussion}.
%
%

The data and code are available at
\begin{small}
\url{www.github.com/anmoisio/type-diversity}.
\end{small}

\section{Grammatical Framework: concrete \& abstract syntax and interpretation} \label{sec:gf}

In this work, we use Grammatical Framework (GF) as a conceptual tool to define type diversity, as well as a practical tool to generate data.
This section is a brief introduction to GF's terminology and syntax.
For further details, see \citet{ranta2004grammatical} for a general introduction, the reference manual\footnote{\url{https://www.grammaticalframework.org/doc/gf-refman.html}} for up-to-date syntax, and \citet{ranta2004computational} for logical semantics in GF. See Appendix~\ref{sec:data-generation-cogs-slog} for further details on our data generation process.


\citet{curry1961some} proposed to consider grammar on two levels: the underlying grammatical structure, called \emph{tectogrammatics}, and the representation of the grammar in terms of concatenating strings, called \emph{phenogrammatics}. Tectogrammatics deals with the categories of language and which categories can be combined with which other categories. Phenogrammatics determines how the categories are realised as strings and how the strings are concatenated.
For example, tectogrammar could include the function \texttt{BUY} that takes as arguments the subject and the object, say \texttt{JOHN} and \texttt{COFFEE}.
Phenogrammar determines the rules how to represent this function as a string: in English the rules say that the order is subject-verb-object, and the verb is inflected to agree with the third-person singular subject: \emph{John buys coffee}.
In computational linguistics, the division into tecto- and phenogrammar guides the design of such frameworks as Abstract Categorial Grammars \citep{de-groote-2001-towards}, as well as GF.

In GF, the tectogrammar is called an \emph{abstract syntax}, and it defines the categories (i.e. types, such as verbs V, nouns N, and sentences S) and the functions that combine these categories. The abstract syntax is defined in GF using the keyword \texttt{cat} for \emph{categories} and \texttt{fun} for \emph{functions}, for example:\footnote{
    Note about reading the functional style for those more familiar with imperative languages: here, for example \texttt{Sentence} can be thought of, in imperative programming style, as a function that takes a \texttt{NP} and a \texttt{VP} as arguments and returns a sentence \texttt{S}. Technically, it is actually a function that takes only one argument of type \texttt{NP} and returns a new function of type \texttt{VP -> S}, which in turn takes one argument of type \texttt{VP} and returns a \texttt{S} (hence we use \texttt{->} between `argument' types as well as before the `value' type: arguments and the value are not really that different). This is called \emph{currying}. We can read a simple grammar, such as the above, in the imperative style, but in general this reading is not applicable, as in our interpretation functions (Appendix~\ref{sec:data-generation-cogs-slog}). There, we have for example a function with a definition
   \texttt{\newline
    fun iVP : VP -> Ind -> Event -> Prop ;\newline
    def iVP (Verbphrase v np) i = iNP np ($\textbackslash$y -> iV2 v i y) ;
    }\newline
   We are not giving the Event argument to the iVP function in the pattern matching in the definition, since it is not needed. The definition would be equivalent if we added it:
   \texttt{\newline def iVP (Verbphrase v np) i e = iNP np ($\textbackslash$y -> iV2 v i y) e ; }}

\begin{small} \begin{small} \begin{verbatim}
abstract Lang = {
    cat S ; NP ; VP ; V2 ; V1 ; V ; N ; Det ;
    fun
        Sentence         : NP -> VP -> S ;
        Verb2phrase      : V2 -> NP -> VP ;
        Verb1phrase      : V1 -> VP ;
        Nounphrase       : Det -> N -> NP ;
        V2Verb           : V -> V2 ;
        V1Verb           : V -> V1 ;
        drink_V          : V ;
        a_Det, the_Det   : Det ;
        man_N, coffee_N  : N ;
}
\end{verbatim} \end{small} \end{small}
where the lines
\begin{verbatim}
    man_N, coffee_N  : N ;
    man_N : N ; coffee_N : N ;
\end{verbatim}
are equal.
We follow the GF conventions of capitalising the names of all categories (e.g. \texttt{NP}), and functions (\texttt{Nounphrase}), except lexical functions (\texttt{man\_N}) which are suffixed with their type.
An example of a syntax tree of type \texttt{S} is:
\begin{small} \begin{verbatim}
Sentence (Nounphrase the_Det man_N)
         (Verb2phrase (V2Verb drink_V) (Nounphrase a_Det coffee_N))
\end{verbatim} \end{small}


The phenogrammatics in GF is called a \emph{concrete syntax}, and it defines linearisation types (\texttt{lincat}, e.g. string \texttt{Str}) for each category, and linearisation rules (\texttt{lin}) for each function in the abstract syntax. A concrete syntax for the above abstract syntax could look like this:
\begin{small} \begin{verbatim}
concrete LangEng of Lang = {
    lincat S, VP, NP, V2, V1, V, N, Det = Str ;
    lin
        Sentence    np  vp  = np ++ vp ;
        Verb2phrase v2  np  = v2 ++ np ;
        Verb1phrase v1      = v1 ;
        Nounphrase  det n   = det ++ n ;
        V2Verb      v       = v ;
        V1Verb      v       = v ;
        drink_V             = "drinks" ;
        a_Det               = "a" ;
        the_Det             = "the" ;
        man_N               = "man" ;
        coffee_N            = "coffee" ;
}
\end{verbatim} \end{small}
where the operator ++ concatenates two strings with a space between them. With these rules, the above example syntax tree is linearised into `the man drinks a coffee'.
In a more realistic grammar, the simple string type \texttt{Str} is replaced by a \emph{record} type that may have multiple \emph{fields}. For example, 
\begin{small} \begin{verbatim}
lincat V = {s : Str ; agreement : Agr} ;
\end{verbatim} \end{small}
means that the linearisation type of V is a record with a field named \texttt{s} of type \texttt{Str} and another field named \texttt{agreement} of type \texttt{Agr}. The value of \texttt{agreement} could be given as an argument to an inflecting function to determine the correct inflected form.



Functions can be declared to be \emph{constructors} by replacing \texttt{fun} with \texttt{data}, which enables using them as constructor patterns in \emph{function definitions} \texttt{def}. Function definitions are needed for \emph{computation}, which in our work is used to interpret syntax trees in logical semantics. See Appendices~\ref{sec:logical-semantics} and~\ref{sec:data-generation-cogs-slog} for further details on how we use GF to generate extended variants of the COGS and SLOG datasets.

\section{Defining compositionality and assessing compositional generalisation} \label{sec:compositionality}


In linguistics, the Principle of Compositionality, or ``Frege's principle'', states that `The meaning of a complex expression is a function of the meanings  of its parts and of the way they are syntactically combined.' The principle was given a precise interpretation by \citet{montague1970universal}, as noted by \citet{partee1997montague}: compositionality requires that there is a homomorphism from the algebra of syntax to the algebra of meaning. Formally, a syntax is defined to include a finite set of primitive expressions of specific categories and recursive rules (with syntactic operations $F$) of the type:
\begin{quote}
    If $\alpha$ is a well-formed expression of category A and $\beta$ is a well-formed  expression of category B, then $\gamma$ is a well-formed expression of category C, where $\gamma$ = $F_i(\alpha, \beta)$.
\end{quote}
There is a homomorphism from syntax to semantics if the semantics parallels the syntax, having a corresponding semantic rule (where $G$ is a semantic operation) for every syntactic rule of the type:
\begin{quote}
    If $\alpha$ is interpreted as $\alpha'$ and $\beta$  is interpreted as $\beta'$, then $\gamma$ is interpreted as $\gamma'$, where $\gamma'$ = $G_k(\alpha', \beta')$.
\end{quote}

In natural language processing, from the principle of compositionality is derived the concept of \emph{compositional generalisation} (CG), which could be narrowly construed to mean that a model learns the mapping from syntax trees to logical semantics from training data, and can therefore interpret novel syntax trees. In practice, the concept of CG is rarely used this narrowly, but instead typically in a more general sense: the homomorphism that is exploited to generalise does not need to be between syntax and semantics per se \citep{mccurdy-etal-2024-toward}. 
Furthermore, compositionality is often merely assumed to be a property of an NLP task, without assessing precisely in what sense the task is in fact compositional.

There are also other definitions of compositionality, in addition to Montague's,
that provide alternative approaches to assessing compositional generalisation.
\citet{keysers2019measuring} proposed a quantifiable, machine learning-oriented definition of `compositionality' and a method to measure how `compositional' a train-test dataset split is, called \emph{distribution-based compositionality assessment}, or DBCA.
It is based on
two values (ranging from 0 to 1): the atom divergence that measures how much the distributions of the \emph{primitive elements} differ between training and test sets, and the compound divergence that measures how much the distributions of the \emph{combinations} of atoms differ. A highly `compositional' data split has a low atom divergence and a high compound divergence.
DBCA is a function of only the input sequences or only the output sequences; in principle, this definition is orthogonal to the traditional definition, since it doesn't concern the input-output mapping. In practice, however, a compositional mapping (in the traditional sense) is assumed when the method is applied by the authors.

It is not clear whether, or under what conditions, Transformers can generalise compositionally. There are two approaches to the question: whether Transformers exhibit compositional behaviour and whether they have compositional representation \citep{mccurdy-etal-2024-toward}.
Compositional behaviour has been found at least somewhat lacking when tested with \emph{diagnostic datasets} such as SCAN \citep{lake2018generalization}, COGS \citep{kim-linzen-2020-cogs,csordas-etal-2021-devil}, or CFQ \citep{keysers2019measuring}, with \emph{targeted syntactic evaluations} \citep{lan2024large},
or by assessing LLM outputs \citep{dziri2023faith}.
Some compositional representations can be found in Transformers \citep{murty2022characterizing,lepori2023break,song2025out}, but it has also been shown that compositional representations do not necessarily lead to compositional behaviour, i.e. they are necessary but not sufficient \citep{xu2022compositional,kobayashi2024can,redhardt2026scaling}.

In this work, we use the SCAN and COGS datasets to assess compositional generalisation in Transformers.
SCAN
consists of sentence pairs in natural language and in a `command' language. The models are trained to translate sentences from natural language to the command language. For example:
\begin{quote}
`look opposite left twice' $\rightsquigarrow$\newline
\texttt{I\_TURN\_LEFT I\_TURN\_LEFT I\_LOOK I\_TURN\_LEFT
I\_TURN\_LEFT I\_LOOK}
\newline\newline
`run and turn around right' $\rightsquigarrow$\newline
\texttt{I\_RUN I\_TURN\_RIGHT I\_TURN\_RIGHT I\_TURN\_RIGHT
I\_TURN\_RIGHT}
\end{quote}
where \ \ `$\rightsquigarrow$' \ \ denotes the seq2seq translation that the Transformers are tasked with. SCAN tests compositional generalisation by leaving certain parts of the data systematically out of the training set. For example, one of the verbs, `jump', appears only as a single-token sentence in the training set but appears in longer sentences in the test set.

COGS is a semantic parsing dataset with a simplified neo-Davidsonian semantics, with a list of presuppositions separated by `;', and after that assertions separated by `AND', and variable names (\texttt{x\_0, x\_1, x\_2, ...} ) that correspond to the word index in the input sentence. It includes multiple different compositional generalisation cases and, important to our work, these can be divided into \emph{lexical} and \emph{structural generalisation} cases. Lexical generalisation is needed when a word is used in a new position. Structural generalisation is needed when a multi-word phrase structure, regardless of the lexemes that populate that structure, is used in a new position in the larger structure of a sentence.
For example, in one of the \emph{lexical} generalisation cases of COGS, the noun `hedgehog' always appears  as the
subject of a sentence in the training set, for example
\begin{quote}
`A hedgehog ate the cake.' $\rightsquigarrow$ \ \ 
\texttt{* cake ( x \_ 4 ) ; hedgehog ( x \_ 1 ) AND eat . agent ( x \_ 2 , x \_ 1 ) AND eat . theme ( x \_ 2 , x \_ 4 )}
\end{quote}
and the test set contains sentences where `hedgehog' is the object (`The baby liked the hedgehog.'). In one of the \emph{structural} generalisation cases, on the other hand,
noun phrases that include a prepositional phrase (PP) appear only as objects in training, e.g. `Noah ate the \textbf{cake on the plate}.', but also as subjects in testing, e.g. `The \textbf{pizza in the oven} burned.' We utilise this structural generalisation task in our experiments below, and will refer to it as the `PP-in-Obj$\rightarrow$PP-in-Subj' task.

\section{Why are structural generalisation tasks hard for seq2seq models?} \label{sec:why-is-structural-hard}


 A few studies have shown that lexical generalisation tasks are not particularly difficult for sequence-to-sequence (seq2seq) models like Transformers, but structural generalisation tasks are \citep{kim-linzen-2020-cogs,yao-koller-2022-structural,li-etal-2023-slog}. In these studies, a few different reasons for this discrepancy are offered.
 In the next subsections we discuss three proposed reasons: that the Transformer model is inherently poor at the task; that the task itself is not `well-posed'; and that the concrete surface properties of the specific datasets are the reason they have been difficult for Transformers. We disagree to differing degrees with each of these characterisations. After this section we will show type diversity to be an important factor that explains much of the previous results. 

\subsection{An architecture hypothesis: Structural generalisation is hard for Transformers}
The first and most commonly assumed reason has been that the seq2seq models, such as Transformers, are inherently poor at structural generalisation. \citet{kim-linzen-2020-cogs} conclude: `composition of structures is more challenging [than lexical generalisation] to both Transformers and LSTMs.' Similarly, \citet{yao-koller-2022-structural} claim in their title that `Structural generalization is hard for sequence-to-sequence models'.


As mentioned in Section~\ref{sec:compositionality}, whether or not Transformers can generalise compositionally is still an open question; whether they can generalise structurally is even more uncertain since it has been studied in only a handful of recent works.
Furthermore, most of these works rely on a single seq2seq task type, namely COGS and its variants \citep{kim-linzen-2020-cogs,yao-koller-2022-structural,li-etal-2023-slog,jabbar-etal-2025-distinguishing}.
Because of the black-box nature of neural nets, these models are not yet understood thoroughly enough to verify or falsify an ability or inability to generalise structurally by studying the inner mechanics of the models.
We argue that it is still premature to blame the Transformer architecture before either obtaining the same result from multiple datasets and task types, or recognising some deficiency in the inner workings of the Transformer model.

\subsection{A perspective: Structural generalisation is not a fair task}

Another explanation offered for the discrepancy is that it is actually too much to ask for the models to solve the structural generalisation test cases.
Already before the COGS dataset was published, \citet{geiger-etal-2019-posing} argued that the kinds of tasks that include what \citet{kim-linzen-2020-cogs} and others have later called `structural generalisation' tasks are not \emph{fair}: the training set is `not sufficient to support the requisite kinds of generalization'.
They define a fair training set as one that provides a \emph{baseline learner} with a complete mapping of every possible intermediate function to its possible arguments. The baseline learner that they define is essentially a recursive lookup table that memorizes every function-argument pair it sees during training.

Recalling the COGS example from the previous section, and the GF example grammar from section~\ref{sec:gf}, an example function would be the constructor \texttt{Sentence} that takes as arguments a subject noun phrase and a verb phrase.
For \citeauthor{geiger-etal-2019-posing}, it is fair to give an intermediate function (e.g. \texttt{Sentence}) a novel argument
that has the same logical form as the function's arguments in training. Since all simple noun phrases share the logical form
(`the \textless noun\textgreater' \ \ $\rightsquigarrow$ \ \ `\texttt{*<noun>(x\_1)}', with the definite article, `a \textless noun\textgreater' \ \ $\rightsquigarrow$ \ \ `\texttt{<noun>(x\_1)'} with indefinite)
it is fair to test the models with samples that have new simple noun phrases in the subject role (lexical generalisation).
But because different linguistic structures have different logical forms, e.g. the prepositional phrase structure
`the \textless noun\textgreater on the \textless noun2\textgreater' has a logical form which is different from that of the simple noun phrases:

\begin{small} \begin{verbatim}
*<noun>(x_1) ; *<noun2>(x_2) ; <noun>.nmod.on(x_1, x_2)
\end{verbatim} \end{small}
it is \emph{not} fair, based on \citeauthor{geiger-etal-2019-posing}, to test the models with samples that have a PP in the subject NP, if the training set did not include this function-argument pair (structural generalisation). 

\citet{wu-etal-2023-recogs} (henceforth WMP) saw that the structural generalisation task PP-in-Obj$\rightarrow$PP-in-Subj in COGS is not fair in this sense, but they suggest (tentatively), as we do (not merely tentatively), that it is still a reasonable task for assessing models. We note that not only this task, but in fact all three of the structural generalisation tasks of COGS are unfair in this sense.
In a recent work, \citet{jabbar-etal-2025-distinguishing} argued in even stronger terms that the definition is about a \emph{necessary} requirement for a fair task. They called the difference between a lexical generalisation task and a structural generalisation task  `fundamental'
and argued that this structural generalisation case (the PP-in-Obj$\rightarrow$PP-in-Subj task) is not `well-posed'.
They backed this claim of ill-posedness
by showing \textbf{(1)} that a lexical generalisation task can be made harder by breaking the distributional similarities between one lexical item (the test case) and others of the same class (their sections 4.2, 4.3, 4.4), \textbf{(2)} that a structural generalisation task becomes easier if you add some of the test set samples into the training set, converting it into a lexical generalisation task (section 4.7), and \textbf{(3)} that pretrained LMs, which have seen the structure but not the specific COGS test samples, have no trouble generalising (lexically) to the samples (section 4.6). However,
none of these results show that structural generalisation is not well-posed, only that \emph{these specific} structural generalisation cases are \emph{harder} than the specific lexical cases.

We argue that the definition of fairness by \citet{geiger-etal-2019-posing} is too strict.
First we note that there seems to be some confusion whether they are defining a \emph{sufficient} or a \emph{necessary} property of a fair training set. In their Definition 2, they state that they define `A Property Sufficient for Fairness', but right after that they call it a `minimal requirement' (and again on page 4491), implying that this is, in fact, a necessary requirement for a fair task instead of claiming that this is merely one sufficient property and there might be also other properties sufficient to make a task fair.
Although we think that the requirement is too strict to be necessary since it would mean we cannot expect structural generalisation from our models, we agree that it would be sufficient for fairness.

An intuitive
way to determine what can reasonably be expected of neural NLP models  is to compare them to humans. 
It is easy to compose structures that readers can understand even though the structures are novel to them. Consider, for example, the sentence:
\emph{
`The fact that the hedgehog ate almost the whole cake and went to sleep just after midnight in the small yet comfortable burrow that she had excavated last summer with great effort impressed Noah.'
}
It is unlikely that you have seen or heard this exact structure as the subject of a sentence before (or if you have, it is easy to extend it until it is novel), but probably you can nevertheless parse and understand the sentence. In other words, you can \emph{generalise structurally}. Since we are concerned with human-like compositional generalisation, it is reasonable to expect structural generalisation from the seq2seq models as well (as argued also by \citet{kim2024structural}): it is fair to expect seq2seq models to process sentences that include a structure in a new role, as in the PP-in-Obj$\rightarrow$PP-in-Subj task.


\subsection{A concrete syntax hypothesis: Difficulty is on the surface} \label{sec:surface}

WMP
provide a third explanation for the discrepancy: the seq2seq models fail because of incidental details of the logical form in COGS. The issue they identify in the syntax is that since a PP is never in a subject NP, a variable with a small number (mainly \texttt{x\_1}) is never bound to PP semantics, and PP semantics is never generated as the first assertion. Seq2seq models learn these kinds of distributional constraints well.

They design three modifications to relax the constraints on the variable numbers and on the order of words: they add filler words such as `um'; they change the word order (preposing), e.g. from `Emma was sold the box in the tent.' to `The box in the tent Emma was sold.'; and they add participial adjectival phrases (PAPs), e.g. `A man \emph{painting the spaceship} froze.', which is given the following semantics: 

\begin{verbatim}
*spaceship(x_4) ; man(x_1) AND man.acl.paint(x_1, x_4) AND
    freeze.theme(x_5, x_1)
\end{verbatim}
where `\texttt{acl}' is presumably short for `Adjectival Clause'.

Filler words and preposing are straightforward ways to test their hypothesis that
the constraints on variable numbers and the order of tokens (both input and output)
make the task unnecessarily difficult.
They show that these modifications increase the accuracy from 0\% to 20.5\%.
This is evidence in favour of their hypothesis, but 20.5\% is still far from the accuracy on the lexical generalisation tasks (in many cases close to 100\%), suggesting that this is not the only difficulty in the task.

However, adding participial phrases is not only a modification of the order of tokens or the variable numbers. Even though they claim that none of these strategies change the set of meanings expressed by COGS, the \texttt{acl} term is a new kind of meaning added to the dataset. Furthermore, they also test a variation where the \texttt{acl} term is changed to the \texttt{nmod} term, i.e. the same semantic form as that of PPs, essentially introducing the participial phrase as a new word in the same category as the existing prepositions. In this mapping, `a man painting a spaceship' maps to the logical form the same way as `a man on a spaceship', `a man beside a spaceship', and `a man in a spaceship', i.e. `\texttt{man(x\_1) AND spaceship(x\_2) AND man.nmod.<prep>(x\_1, x\_2)}', where \textless prep\textgreater is replaced by `in', `on', `beside', or 
`painting'. (Note that in the natural language sentences, `painting' and `painted' have no relation as far as the models are concerned, since the convention for COGS has been to use only full words as tokens, instead of breaking them into morphemes `paint' + `ing'.)
With this addition to the training set, the structural generalisation case is reduced to a lexical generalisation case (similarly to sections 4.6 and 4.7 of \citet{jabbar-etal-2025-distinguishing}), where the training set does \emph{not} include subject NPs containing a prepositional phrase with `in', `on', or `beside', but it \emph{does} include subject NPs containing a participial phrase, which have identical semantics with PPs in this setup. With this modification, the accuracy of transformers is increased from 0\% to 82.7\% in the PP-in-Obj$\rightarrow$PP-in-Subj task.

Using the \texttt{acl} token for participial phrases, instead of \texttt{nmod}, increases the accuracy from 0\% to 24.7\% in the PP-in-Obj$\rightarrow$PP-in-Subj task. This setup does not reduce the task to lexical generalisation, but it does introduce a new type of structure into the natural language and a corresponding meaning into the semantics. If we wanted to test
their hypothesis that the constraints on token order and variable numbers make the task difficult,
we would add these only in the subject, i.e. in the position that is filled by a prepositional phrase in the test set. But instead they add them to both subject and object positions.
We suspect that the performance increase from 0\% to 24.7\% is not actually caused by the changes in the index numbers and word order, which we test in the next section.

\section{An abstract syntax hypothesis: \emph{Type diversity} enables generalisation} \label{sec:hypothesis}

\subsection{A preliminary experiment}  \label{sec:preliminary}

To test our hunch from the previous section, we create a new training set variant by removing the participial phrases from the objects, leaving them only in subjects. 
Table~\ref{tab:prelim-exp} shows the replicated numbers for the experiment by
WMP
and  the numbers for the data variants where PAPs are only in subjects of sentences.
If the performance improvement is due to the word order, having the PAPs only in subject should have exactly the same effect as having them in both subject and object positions. However, what we find is that the PP-in-Subj accuracy decreases back close to 0\%, which cannot be explained by the reasoning of
WMP.

\begin{table}[h]
\centering
\small
\begin{tabular}{lcc}
    \toprule
    Training dataset variant & Semantics & Accuracy on PP-in-Subj (\%) \\
    \midrule
    Original COGS  &      PP=\texttt{nmod} (no PAPs)           & 0.0 \ \ ($\pm$0.0) \\ 
    PAPs in subject and object &PAP=\texttt{acl}, PP=\texttt{nmod}                     & 32.0 \ \ ($\pm$17.6) \\ 
    PAPs in subject& PAP=\texttt{acl}, PP=\texttt{nmod}                                & 0.2 \ \ ($\pm$0.4) \\ 
    PAPs in subject and object& PAP=PP=\texttt{nmod}             & 83.8 \ \ ($\pm$1.5) \\ 
    PAPs in subject &PAP=PP=\texttt{nmod}                       & 81.8 \ \ ($\pm$4.0) \\ 
    \bottomrule
\end{tabular}
\caption{Transformer accuracies in the PP-in-Obj$\rightarrow$PP-in-Subj task of COGS for different variants of the COGS training dataset. }  \label{tab:prelim-exp}
\end{table}

Interestingly, in the case where the structural generalisation task is degenerated into a lexical generalisation task by using the same semantics for PAPs and PPs, Transformers are helped almost as much by having PAPs in only subject. This verifies that the clues these two modifications, with \texttt{nmod} or \texttt{acl} semantics, offer to the models about the generalisation target are two different ones. The \texttt{nmod} semantics in practice reduces the PAP into a PP, giving the models directly the information that PPs can appear in the subject, which only requires that this pseudo-PP appears in the subject. 
The \texttt{acl} semantics shows the models only that subjects can include PAPs too: if NPs with PAPs never appear in the same position (object) as NPs with PPs, it is a greater leap to infer that NPs with PPs and NPs with PAPs are of the same type (NP) at all.

\subsection{Type diversity}  \label{sec:type-diversity}
Where does the accuracy gain from 0\% to 24.7\% (or 32.0\% in our replication of the experiment) come from, if not the changes in variable names and order of tokens? We argue that the most significant change introduced by the PAPs (with the \texttt{acl} token as the semantics) is that \textbf{a new NP constructor is added}, which makes it easier to infer that there is a type NP, meaning all trees of type NP are interchangeable without breaking grammaticality and they map to semantics the same way.

\citet{patel-etal-2022-revisiting} showed that simply adding more words in the lexicon, i.e. increasing \emph{lexical diversity}, makes lexical generalisation easy for seq2seq models.
Specifically, they added words \emph{of the same type} as the test word, namely verbs.
%
We reason that, analogously, adding more structures of the same type helps generalising to a new structure, and, generally, \textbf{adding to the training set more constructors of the same type as the test case helps generalising to it}.
We call this \emph{\textbf{type diversity}} (TD) to emphasise that it is not lexical or structural diversity in general, but type-specific diversity.
The functional paradigm of GF helps us see that TD applies both to lexemes and to multi-word structures, since they are both \emph{constructor functions} and treated equally in the abstract syntax of GF.
However, we highlight that when the diversity of structural types is increased, new structures of the same type are added instead of familiar structures populated by new words.

Concretely, \citet{patel-etal-2022-revisiting} added to the training set more constructors of type V, i.e. they increased TD of the type \texttt{V} by adding new verbs, which helped generalising to a specific verb in a new position. Similarly, adding to the training set new NP constructors that include, for example, participial phrases, adjectives (`a lazy man'),
or relative clauses (`a man that painted a spaceship'), which appear as both subjects and objects, should help generalising to the target structure of NP with PP in the subject role.

We could also call this `filler diversity' or `role diversity', since `type' here refers to the category of structures that fills a specific role: an NP fills the subject or object role, and this is the relevant fact with regards to generalisation in this test case that we have discussed.
However, we call it `type diversity' to emphasise that in practice we quantify it by counting the number of constructors the type has in the specific grammar we use.

This means type diversity depends on the grammar; in this work we use the types defined in GF. We mostly focus on uncontroversial types such as noun phrases: it seems obvious that such phrases as `a cat',  `a cat on a mat', and `a big cat' belong to the same category, and there are no reasonable alternative grammars that place them in different categories. However, how these NPs are built may differ: in CCG \citep{steedman2000syntactic} a post-nominal PP is typed \texttt{(NP\textbackslash NP)/NP}, so on `a mat' combines with the already-determined NP `a cat' \citep{hockenmaier-steedman-2007-ccgbank}, but in GF the PP attaches to the common noun (CN) `cat' without the determiner, which is the canonical way \citep[ch. 4]{montague1973proper,barwise1981generalized,Heim1998HEISIG}.
What this means for our experiments is that the constructor that builds the phrase `cat on a mat' is not actually an NP constructor but a CN constructor, as is the PAP constructor.
However, 
in our datasets, trees of type CN never appear outside an NP, so every new CN constructor is also a new NP structure; we can therefore ignore CN diversity and count it as NP diversity
(cf. compound weights in \citet{keysers2019measuring}). 

\subsection{Previous work on the effects of dataset properties on compositional generalisation}

The effects of training data properties on compositional generalisation have been studied from at least two different perspectives with two different goals: aiming to develop data augmentation methods that help improve performance, or simply trying to understand the effects that data properties have on model performance, of which the latter is our goal. We first note some differences between these perspectives and then review previous works.

The \emph{data augmentation} research frames training dataset modifications as a class of methods that help achieve better results, comparable to modifications of model architecture or learning algorithm.
An assumed restriction in data augmentation is that the grammar or other generative function that produces the data is not available to the model. Our goal is not to develop a data augmentation method
so we do not restrict our experiments (below) in this way.
The most significant differences between our work and data augmentation research are that in data augmentation (1) the aim is to increase the training set size and (2) if a data augmentation method is able to recover some of the test samples, they can be used in training.
Therefore, this line of work conflates increasing type diversity (i.e. adding new constructors of the same type as the test case) with simply adding exactly the same constructor as the test case into the training set. 
In other words, a data augmentation method is performing well if it can recover test samples so that the model \emph{does not need to} generalise, whereas we aim to find the data properties that \emph{enable} the models to generalise.
In our experiments, the training set sizes are kept approximately equal, and we never include the test cases in training.

\subsubsection{Data augmentation} \label{sec:data-augmentation}
 
Multiple data augmentation studies have devised clever ways to synthesise lexical diversity, e.g. more verbs in SCAN, reporting significant gains especially in the add jump split, similarly to the aforementioned study by \citet{patel-etal-2022-revisiting}. \citet{jiang-etal-2022-mutual} proposed primitive mutation, i.e. generating new versions of the verbs, e.g. `walk left twice' is mutated to `walk2 left twice' $\rightsquigarrow$ \texttt{I\_TURN\_LEFT WALK2 I\_TURN\_LEFT WALK2}, and \citet{zhou-etal-2023-data} extended this method to all tokens `walk2 left2 twice2' $\rightsquigarrow$ \texttt{I\_TURN\_LEFT2 WALK2 I\_TURN\_LEFT2 WALK2}.
The GECA method \citep{andreas-2020-good} `aims to discover substitutable sentence fragments [...], with the fact that a pair of fragments appear in some common sub-sentential environment [...] taken as evidence that the fragments belong to a common category'. The fragments are either one token or two consecutive tokens, which means they can't discover for example the noun phrases with prepositional phrases `a cat on a mat' as instances of noun phrases in COGS. The examples of fragments they report for SCAN are all verbs such as `jump'. LexSym \citep{akyurek-andreas-2023-lexsym} is a similar method that induces a lexicon in order to synthesise new sentences by swapping word-semantics pairs (i.e. input token aligned with its corresponding output token) with others in the same category. As the name suggests, the method categorises only lexemes, not multi-word structures like noun phrases.

The methods by \citet{qiu-etal-2022-improving,yang-etal-2022-subs} augment the training data by substituting multi-word subtrees in samples, making them in this sense more general and closer to targeting type diversity. In practice, however, in their experiments their methods are based on recovering the test samples and including them in the training set, instead of creating diverse new samples that help generalising beyond the training set. Similarly, \citet{yao-koller-2024-simple} show that sampling output sequences from the \emph{test} distribution and backtranslating new input sequences for them
helps when evaluating on the test set.

Though most of these data augmentation methods could be partly, implicitly based on increasing type diversity (e.g. adding a new verb `walk2' in the training set that acts similarly to `walk'), and some of them allude to it obliquely (e.g. aiming to find a common category), none of them spell out the same linguistically-motivated and general definition as we did in Section~\ref{sec:type-diversity}, and none of them aim to increase it specifically.


\subsubsection{Understanding the effects of data properties}

We already mentioned the work by \citet{patel-etal-2022-revisiting}, who added verbs in SCAN and found that seq2seq models became able to generalise to the held-out verb `jump'.
\citet{11093806} define \emph{diversity} to mean the number of values an attribute can take in a visual recognition task, namely the colour of an object in the CLEVR dataset \citet{johnson2017clevr} in which the task is to recognise visual attributes of objects in an image. In this visual parallel to adding verbs to SCAN, when the colour attribute is more diverse the models generalise better to new combinations of attributes.
Similar work include that by
\citet{pmlr-v235-ramesh24a}, who showed that which compositions appear in training determines which unseen ones are solvable.


\emph{In-context learning} (ICL) \citep{brown2020language} has been the focus of much of the recent work on generalisation.
\citet{shin-etal-2022-effect} showed that the training set source, not just size, can determine whether ICL appears. 
\citet{chan2022data} reported how the data properties burstiness, Zipfian long tail of rare classes, synonyms, and homonyms help Transformers in ICL. Of these properties, \citet{nallamala2024mechanistic} replicates the first three in a two-parameter model and shows that they are caused by induction-head formation in this minimal network.
\citet{raventos2023pretraining} defined \emph{task diversity} as the number of tasks in the pre-training, and found a threshold diversity above which ICL is able to generalise. \citet{Goddard2025when} add a second dimension to this analysis. 

\emph{Scaling} \citep{kaplan2020scaling} is another closely related topic, and recent work has demonstrated the importance of data properties besides training set size.
\citet{chen-etal-2025-revisiting} propose a redundancy metric combining the concentration of samples within individual clusters and the separation between different clusters, where denser means more redundant, less diverse. They find that training on dense data hinders learning.

The density metric is directionally similar to what we call `type diversity', but aims to be more general, applying across machine learning tasks, whereas type diversity is defined in linguistic terms, applying only to language tasks. Focusing on language offers a different perspective compared to the data-agnostic metrics such as density. Language has an intricate structure that has been studied for millennia; type diversity as a metric leans on this previous work of (computational) linguists that has delineated the categories (or types) present in language, and focuses on one specific kind of diversity instead of lumping all diversity together.


\citet{misra-mahowald-2024-language} found that when a \emph{construction} \citep{goldberg2019explain}  is absent from the training data, namely the AANN construction (`a beautiful five days'), other related factors help the models generalise to it. The related factors include similar NPs in the training data, such as DT ANNs (`the beautiful five days). In principle, this is a specific case of type diversity helping the models to generalise; in practice, it seems more important that the two constructions are very similar.



\section{Using GF to generate and extend the SCAN, COGS, and SLOG datasets} \label{sec:data}

The natural-language-to-command-language dataset SCAN can be  generated by a quasi-synchronous context-free grammar (QCFG) \citep{smith-eisner-2006-quasi},
a variant of the \emph{syntax directed translation scheme} \citep{Aho1969Properties,Aho1969Syntax,wu1997stochastic} that allows repetition unlike the original SDTS. Repetition is needed e.g. by the `twice' function of SCAN.
The SCAN grammar is simple; see
Appendix~\ref{sec:data-generation-scan}
for SCAN written in GF.

The semantic parsing dataset COGS and its extension SLOG, on the other hand, are more interesting datasets to generate since, as noted by \citet{qiu-etal-2022-improving}, they require variable binding and cannot be generated using a QCFG.
In previous works that generate COGS with a QCFG, variable binding is accomplished by a post-processing step that leans on the order of the words that the variables refer to \citep{kim-linzen-2020-cogs, qiu-etal-2022-improving, li-etal-2023-slog}. This makes the data generation process brittle and difficult to extend to new kinds of sentence-semantics samples.

GF offers two important advantages over QCFGs to generating semantic parsing datasets.
Firstly, the fragment of English used in COGS is included in the Resource Grammar Library (RGL) of GF, from which we can simply select the needed fragment of language. We therefore need to actually write a new grammar only for the logical semantics side of the dataset. In fact, the RGL includes grammars for a much larger fragment of English (and for many other languages too), making it easy to extend the datasets with more linguistic diversity.
Secondly, in GF, variable binding is possible together with the interpretation to logical semantics using \emph{higher-order abstract syntax} \citep{ranta2004computational}, which guarantees that the seq2seq data generation process is compositional and makes it more robust and extendable to new kinds of linguistic structures. We make use of these features when creating linguistically diversified variants of COGS in Sections~\ref{sec:expts-cogs} and ~\ref{sec:expts-slog}. See Appendix~\ref{sec:data-generation-cogs-slog}
for details on the data generation process.

The logical semantics format that we use is slightly different from the original COGS format.  The main differences between the original format and ours are:
    (1) Similarly to WMP, 
    we separate event predicates and thematic role predicates, and rely on variable binding to link them: e.g., change `\texttt{agent.eat(6,7)}' to `\texttt{eat(6) AND Agent(6,7)}'.
    (2) The variable names come from GF: \texttt{v0, v1, v2, ...} instead of \texttt{x\_0, x\_1, x\_2, ...}.
    (3) We add predicates for the morphological features of tense (anteriority and time) and number.
    (4) We re-tokenise the sentences into
    subword tokens that approximate
    morphemes, e.g. `discover ed by the scientist s', which aids morphological generalisation.
    (5) The order of the semantic terms is determined by our interpretation functions instead of the order of the words in the natural language sentence.
We assess the effect of these modifications in Section~\ref{sec:expts-slog}.

\section{Experiments on structural generalisation and type diversity in COGS} \label{sec:expts-cogs}

In this section we continue to examine Transformer generalisation in the PP-in-Obj$\rightarrow$PP-in-Subj task of COGS.
In addition to participial adjectival phrases (PAPs) as in
WMP,
we increase NP diversity by adding adjectives, relative clauses, and plural forms of nouns into the training set.
We also add new verb tenses, which are different from the other additions in that they do not add diversity to NPs. Our hypothesis predicts that they should therefore not help in this task.
We train Transformer models using the hyperparameters tuned by
WMP,
using 12 random seeds, and report the mean accuracy and standard deviation across seeds. We measure accuracy by checking that the sets of assertions and presuppositions match those of the correct output: the order of the sequence does not matter, as the logical form is a flat conjunction.

Table~\ref{tab:cogs-results} lists the accuracy in the COGS PP-in-Obj$\rightarrow$PP-in-Subj
task.\footnote{COGS has two other structural generalisation cases, but these two are rendered practically impossible to solve by seq2seq models since they add new variable names to the test set, posing an out-of-vocabulary problem, as noted by
WMP.
\label{fn:cogs-struct}}
The first line is the original COGS training set that includes three CN or NP constructors (CN are essentially NP constructors as noted in Section~\ref{sec:type-diversity}): simple common nouns `a cat', NPs with PPs `a cat on a mat' (only in object role), and proper nouns (`Alice'). However, the output sequence format is not exactly the same (see Section~\ref{sec:data} for the differences), which raises the accuracy from 0.0\% in the original to 2.1\% here.

Each addition to the diversity increases the dataset size (from 1656195 tokens, including input and output, in the original dataset to 2655262 in the dataset with all additions).
We make sure this does not confound the results by adding 60\%  more samples to the original COGS training dataset, matching the word count of the largest modified dataset in this experiment, i.e. the bottom row in the table. We create new samples by taking original sentences and swapping their words (within word class) to new ones; e.g. `Mary saw a cat on a mat.' can be mutated into `Emma painted a cup on a table.' This data augmentation makes essentially no difference: the accuracy is changed from 2.1 to 1.7\%, verifying that dataset size does not confound the results.

\begin{table}[htb]
\small
\begin{tblr}{
  colspec={@{}X[halign=l,1.3]X[halign=l,1.8]X[halign=c,1.25]@{}},
  column{1,2}=,
  row{1}={font=\bfseries},
  hline{1,2,Z}={solid},
}
\newline Dataset variant    & \newline Example sentence                                             & Transformer accuracy for PPs-in-subject (\% $\pm$std) \\
Simple NPs + PPs in object  & Mary saw a cat on a mat.                                              & 1.7  ($\pm$1.7) \\    
\ \ \ + PAPs in subject     & A dog eating a bone saw a cat on a mat.                               & 6.2  ($\pm$4.3) \\    
\ \ \ + PAPs in object      & A dog eating a bone saw a cat chasing a mouse.                        & 16.8 ($\pm$8.5) \\    
\ \ \ + adjectives          & A big dog eating a bone saw a cat chasing a small mouse.              & 26.0 ($\pm$13.7) \\   
\ \ \ + RCs                 & A big dog eating a bone saw a cat chasing a small mouse that walked.  & 55.4 ($\pm$18.2) \\   
\ \ \ + plural              & Big dogs eating a bone saw cats chasing a small mouse that walked.    & 69.3 ($\pm$12.7) \\   
\ \ \ + tenses              & Big dogs eating a bone see cats chasing a small mouse that will walk. & 73.3 ($\pm$10.2) \\   
\ \ \ + artificial NPs      & Big dogs a1 a bone see cats chasing a small mouse that will walk.     & 92.4 ($\pm$1.9) \\    

\end{tblr}
\caption{Accuracies in the PP-in-Obj$\rightarrow$PP-in-Subj task of COGS for different variants of the COGS training dataset. Each row includes also the additions of the rows above it.}  \label{tab:cogs-results}
\end{table}

The datasets at rows `+ PAPs in object' and `+ PAPs in subject' are in principle the same as in
our preliminary experiments in Section~\ref{sec:preliminary}.
Again, the difference between 0.2/32.0\% in Section~\ref{sec:preliminary} and  6.2/16.8\% here comes mainly from the differences in the output format as well as the variation across random seeds.
Unlike in the preliminary results, there is some improvement when we add PAPs only in the subject (from 1.7 to 6.2\%), which confirms that the surface properties do have some effect on the results too (see Section~\ref{sec:expts-slog} for more discussion and experiments on the surface properties). Still, a much larger improvement comes from adding PAPs also in the \emph{object} role; recall that the test case includes PPs in \emph{subject}.
Similarly to PAPs, including adjectives and relative clauses in CNs both improve the results significantly, together up to 55.4\%.

Next we add morphological diversity. Like the previous additions, plural forms of nouns add diversity to noun phrases, although this time via the type \texttt{Det}.
Plurals add a new constructor of type \texttt{Num}, which is an argument of the \texttt{DetQuant : Quant -> Num -> Det} constructor that combines the quantifier (`a'/`the') and the number (singular/plural).
Simple common noun NPs are built from \texttt{Det} and \texttt{CN} by the constructor \texttt{DetCN : Det -> CN -> NP}.
As with \texttt{CN}, trees of type \texttt{Det} never appear outside NPs, so we can collapse the distinction: the diversity of the \texttt{CN}, \texttt{Det} and \texttt{Num} types is in practice NP diversity, in our dataset. 
Again, the performance improves significantly, from 55.4 to 69.3\%.

We also add different verb tenses:
\emph{time} can be, in addition to \emph{past} in original COGS, \emph{present}, \emph{future} or \emph{conditional}; and \emph{anteriority} can be \emph{simultaneous}, as in all the original samples, or \emph{anterior}. These two dimensions add seven more tenses to the single past+simul tense in COGS, in total eight: \emph{walks, walked, will walk, would walk, has walked, had walked, will have walked} and \emph{would have walked}.
This addition does not add NP diversity, and
therefore it is congruent with our hypothesis that this addition does \emph{not} improve the performance significantly.
The mean accuracy is somewhat higher than without this addition (73.3 vs 69.3\%) but this is less than half of the standard deviation across different random seeds.

There is a robust correlation between type diversity and performance, with each addition of type diversity improving the performance more, up to around 70\%.
Finally, to show that this trajectory of the mean accuracy driven by type diversity does not plateau before close to 100\%, we add two artificial NP structures of the form:
`the dog a2 a deck' \ \ $\rightsquigarrow$ \ \
\texttt{*dog(v1); deck(v2) AND PrepArtif2.a2(v1, v2)}.
Both of the structures (\texttt{PrepArtif1, PrepArtif2}) include three nonce words, `a1'/`b1'/`c1' and `a2'/`b2'/`c2'.
These structures are different from the test case, unlike in the experiment by
WMP
where the semantics had the same token `nmod' as the test case; in our experiment the models still need to generalise structurally instead of lexically, as in the `acl' variant of the experiment by WMP.
This addition brings the accuracy up to 92.4\%, showing that accuracy correlates with type diversity all the way from nearly zero to above 90\% in this structural generalisation task.

A natural  question to ask  is, do the diversities of structural and lexical types \emph{in real datasets} make structural generalisation harder than lexical?
Since the structural generalisation result, before the artificial NPs, was only around 70\% instead of 100\% as in some lexical generalisation cases, it's conceivable that structural generalisation is somewhat harder than lexical in real NLP tasks if the diversities of structural types are in fact much lower than those of the lexical types in real datasets.
Nevertheless, it seems more plausible to us that the gap between structural and lexical generalisation is insignificant when Transformers are trained on real datasets.
In our highest-diversity dataset, there are five NP or CN structures (simple common noun, proper noun, PP, PAP, and RC) but this is not close to exhausting all NP structures that appear in English. To mention a few, there are mass nouns, genitive/possessive (`the cat's owner', `the owner of the cat'), compound nouns (`a fire engine'), appositive (`my friend Bob'), nominalised gerund (`running' in `running is fun'), etc. Real quantifiers also include others besides `a'/`an' and `the' which we use: for example `every', `no', `this', and `that'. Writing the interpretation functions for all of these would have been a significant amount of work, so we settled for the five natural constructors + the artificial. We argue that the improvement that each added constructor brings to the Transformer accuracy provides enough evidence to conclude that the difficulties of Transformers in structural generalisation that previous work have reported are due to the discrepancy of lexical and structural type diversities. 
In COGS, the diversity of the lexical types range from 20 (verbs that take complement sentences such as `say') to about 400 (nouns).





In summary, besides the strong correlation between TD and accuracy, we (1) noticed that the concrete properties do have an impact on the results too, as shown also by previous work 
by WMP,
and (2) described how defining TD is complicated by the diversities of closely related types, such as \texttt{Det} and \texttt{CN} in relation to \texttt{NP}. In the next section, we map more comprehensively the effects of type diversity and \emph{sub}- and \emph{supertree} type diversity.
(We
call T a `supertree' of S when S is a
subtree of T.)
After that, we investigate the effects of the more superficial factors in Section~\ref{sec:expts-slog}.

\section{Experiments in the SCAN domain}
\subsection{Structural generalisation and (sub/supertree) type diversity in SCAN} \label{sec:expts-scan}

 
Section~\ref{sec:expts-cogs} showed how adding type diversity can increase Transformer accuracy in the structural generalisation task of COGS from around to 2\% to above 90\%.
In this section we further map the relationship of the target type diversity and the diversity of related types with generalisation performance in the natural-language-to-command-language task of the SCAN dataset \citep{lake2018generalization}. This simpler domain enables controlling the degree of diversity more comprehensively.

\subsubsection{Experimental setup}
Original SCAN does not include structural generalisation cases, but these are easily added.
We investigate how type diversity affects three types that are in hierarchical relation to each other, similarly to how, for example, verbs, verb phrases, and clauses appear in natural language: a clause includes a verb phrase, which in turn includes a verb.
For each of the three types, we hold out one constructor for the test set to assess generalisation, and add different numbers of other constructors of this type in new variants of SCAN to assess the relationship of type diversity and compositional generalisation. Table~\ref{tab:scan-variants} summarises the three types.

\begin{table}[htb]
\small
\begin{tblr}{
  colspec={@{}cX[halign=l,1.7]X[halign=l]X[halign=l,1.7]@{}},
  column{1,2}={font=\ttfamily},
  row{1}={font=\bfseries},
  hline{1,2,4,6,Z}={solid},
  hline{3,5}={2-4}{solid},
}
  Type & Example tree & Input & Output \\ 
  V & jump\_V & jump  & I\_JUMP \\
   & pbexl\_V & pbexl &  I\_PBEXL\\
  VP    & OppositeVP((VVerb jump\_V), right\_Adv) & jump opposite right & I\_TURN\_RIGHT I\_TURN\_RIGHT I\_JUMP \\ 
        & AdvVerbAdv((VVerb pbexl\_V) left\_Adv)) & pbexl ava left & I\_TURN\_LEFT I\_PBEXL I\_TURN\_LEFT \\
  Imp & VP2VP3VP1 \textcolor{blue}{AdvVerbAdv((VVerb pbexl\_V) left\_Adv))} \textcolor{red}{(UseV run\_V)} \textcolor{green}{(OppositeVP((VVerb jump\_V), right\_Adv))} & vp2vp3vp1 \textcolor{blue}{pbexl ava left} \textcolor{red}{run} \textcolor{green}{jump opposite right} & \textcolor{red}{I\_RUN} AND \textcolor{green}{I\_TURN\_RIGHT I\_TURN\_RIGHT I\_JUMP} AND \textcolor{blue}{I\_TURN\_LEFT I\_PBEXL I\_TURN\_LEFT} \\
\end{tblr}
\caption{Example trees and sequences of the three types whose constructors the models need to generalise to and whose diversity is increased. See  Appendix~\ref{sec:data-generation-scan}
for the SCAN grammar.}\label{tab:scan-variants}
\end{table}

The type V is the lexical type: each constructor is one lexicon entry, and generalising to a new constructor is lexical generalisation. Originally, the type V has four constructors for the verbs `walk', `run', `jump', and `look'. We hold out `jump' by including it only as the primitive in the training set (`jump'\ \ $\rightsquigarrow$ \ \ \texttt{I\_JUMP}), similarly to the original SCAN `add jump' split. In addition to the three other verbs that appear freely in the training set, we add  0, 10, 30, 60, or 90 new V constructors for random nonsense words that act like verbs, e.g. `qwert left twice'\ \ $\rightsquigarrow$ \ \ \texttt{I\_TURN\_LEFT I\_QWERT I\_TURN\_LEFT I\_QWERT}, similarly to \citet{patel-etal-2022-revisiting}.

The second type, VP, originally has the constructors for the `opposite' and `around' structures, as well as for a simple verb and for a verb with a direction, e.g. `run left'. The abstract syntax functions for the VP constructors are:

\begin{small} \begin{verbatim}
data UseV : V -> VP ;
data DirVP, OppositeVP, AroundVP : Verb -> Adv -> VP ;
\end{verbatim} \end{small}
The input side concrete syntax is:
\begin{small} \begin{verbatim}
lin UseV       v     = v ;
lin DirVP      v adv = v ++ adv ;
lin OppositeVP v adv = v ++ "opposite" ++ adv ;
lin AroundVP   v adv = v ++ "around" ++ adv ;
\end{verbatim} \end{small}
And the output side concrete syntax is:
\begin{small} \begin{verbatim}
lin UseV       v   = v ;
lin DirVP      v adv = adv ++ v ;
lin OppositeVP v adv = adv ++ adv ++ v ;
lin AroundVP   v adv = adv ++ v ++ adv ++ v ++ adv ++ v ++ adv ++ v ;
\end{verbatim} \end{small}
The type \texttt{Verb} combines the aforementioned type \texttt{V} and the type \texttt{VD} of `turn' that behaves differently than the other four verbs (see Appendix~\ref{sec:data-generation-scan} for the full grammar of SCAN).
The \texttt{UseV} constructor simply creates a VP from a Verb so that it can be used the same way as the more complex VPs. The \texttt{DirVP}, \texttt{OppositeVP} and \texttt{AroundVP} constructors combine and repeat the verb and the direction adverb in a specific order in the output sequences of the dataset. We hold out the \texttt{OppositeVP} constructor by leaving only one introductory sample in the training set: `walk opposite right' \ \ $\rightsquigarrow$ \ \ \texttt{I\_TURN\_RIGHT I\_TURN\_RIGHT I\_WALK}.
The test set includes `opposite' in longer sequences, such as `walk around left and run opposite right', requiring structural generalisation. We then create new constructors that combine and repeat a verb and an adverb in some specific order in the output sequence, e.g.:

\begin{small} \begin{verbatim}
data AdvVerbAdvAdv            : Verb -> Adv -> VP ;
data AdvAdvAdvVerbAdvVerbVerb : Verb -> Adv -> VP ;
\end{verbatim} \end{small}
with concrete syntax linearisation rules, for input side:
\begin{small} \begin{verbatim}
lin AdvVerbAdvAdv            verb adv = verb ++ "avaa" ++ adv ;
lin AdvAdvAdvVerbAdvVerbVerb verb adv = verb ++ "aaavavv" ++ adv ;
\end{verbatim} \end{small}
and output:
\begin{small} \begin{verbatim}
lin AdvVerbAdvAdv            verb adv = adv ++ verb ++ adv ++ adv ;
lin AdvAdvAdvVerbAdvVerbVerb verb adv = adv ++ adv ++ adv ++ verb ++
                                        adv ++ verb ++ verb ;
\end{verbatim} \end{small}
For example, an input-output pair in this new SCAN variant could be
\begin{small} \begin{verbatim}
walk avaa left and look
I_TURN_LEFT I_WALK I_TURN_LEFT I_TURN_LEFT I_LOOK
\end{verbatim} \end{small}

As with all words in SCAN, the nonce words such as `avaa' are not split into letters by the model's tokeniser, so the input word does not contain any clues of the output sequence for the model; the token could be anything, but we define it to contain the order of verbs and adverbs for the convenience of the human reader. We add 0, 1, 10, 30, 60, or 90 more VP constructors in the training set.

The third and hierarchically highest type is Imp (short for `imperative', named nearly arbitrarily since these are nonce words and clauses) and it replaces the `twice' and `thrice' structures in SCAN. The constructors repeat and/or reorder three VPs in the output sequence. The base setup with lowest Imp diversity has two Imp constructors, with the following abstract syntax functions:

\begin{small} \begin{verbatim}
fun VP1VP2VP1 : VP -> VP -> VP -> Imp ;
fun VP3VP2VP1 : VP -> VP -> VP -> Imp ;
\end{verbatim} \end{small}
input side concrete syntax:
\begin{small} \begin{verbatim}
lin VP1VP2VP1 vp1 vp2 vp3 = "vp1vp2vp1" ++ vp1 ++ vp2 ++ vp3 ;
lin VP3VP2VP1 vp1 vp2 vp3 = "vp3vp2vp1" ++ vp1 ++ vp2 ++ vp3 ;
\end{verbatim} \end{small}
and output side concrete syntax:
\begin{small} \begin{verbatim}
VP1VP2VP1 vp1 vp2 vp3 = vp1 ++ "AND" ++ vp2 ++ "AND" ++ vp1 ;
VP3VP2VP1 vp1 vp2 vp3 = vp3 ++ "AND" ++ vp2 ++ "AND" ++ vp1 ;
\end{verbatim} \end{small}

Again, we leave one constructor (\texttt{VP2VP3VP1}) as the test case, including only one simple sample of it in the training set: `vp2vp3vp1 walk look run'\ \ $\rightsquigarrow$ \ \ \texttt{I\_LOOK AND I\_RUN AND I\_WALK}
and add the more complex sequences containing \texttt{VP2VP3VP1} in the test set, for example
`vp2vp3vp1 walk around left turn right look and vp1vp2vp1 look walk turn around left'.
For Imp there are only $3^3=27$ different possible constructors, so we add 0, 4, 8, or 16 new constructors to the training set, besides \texttt{VP1VP2VP1} and \texttt{VP3VP2VP1} which appear in all training sets.

The structural constructors here, of types VP and Imp, are different from the NP-with-PP constructor in COGS in the sense that each of them has exactly one indicator token (e.g. `avaa' or `vp1vp2vp1') in the input sequence, whereas PPs have three different ones: `in', `on' and `beside'. Another difference in the setup is that in COGS the PPs appear frequently in other roles except subject, whereas here we include the test constructor only in one role, namely the simplest role of appearing alone in the sequence.
Structural generalisation is an abstract concept that allows for multiple instantiations: in the other sections we aim to replicate the COGS experiment.
In this section, instead of replicating COGS, we aim to  compare lexical and structural generalisation in a more controlled experiment.

The three types comprise a 3-level hierarchy of structures: Vs are subtrees of VPs, which in turn are subtrees of Imp trees.
Each level has its own generalisation cases in the test set. We could generate a sample that requires generalising in three different ways at once, but we focus on samples that require generalising in only one way in order to distinguish the effects of each of them.
We generate training sets that have an increasing number of constructors for each of the three types.
The dataset sizes are approximately equal, so that the level of type diversity is the main difference between the datasets. The lengths of the training sets, including both input and output sequences, vary between 440,520 and 467,840 tokens,
with an outlier at 388,263 which is the training set with no added constructors for any of the types.
The outlier is due to the data generation process that used the base dataset as a template.
We also control for the sequence length: the average lengths of output sequences vary only between 17.1 and 18.0 tokens for the datasets, again with  the base dataset as an outlier that averages 15.2.
This leaves a possible confounding factor in the base dataset (where V, VP, and Imp diversities are 0), and we do not draw any conclusions from this outlier dataset alone (which is only one dataset out of 120).

\subsubsection{Transformer training details}
There are 5x6x4=120 different training sets, and we train models with 14 different random seeds for each training set and report the mean and standard deviation. The test set is always the same, divided into three parts that test generalising to a new constructor of one of the three types, respectively.

We train Transformer models using mostly the same hyperparameters as \citet{csordas-etal-2021-devil}, with roughly 1M learnable parameters per model.
We use relative positional encodings in the encoder and decoder self-attention. Our variant is simpler than the Transformer-XL–style relative attention of  \citet{csordas-etal-2021-devil}: we add a learned per-head scalar bias, computed as a linear projection of a sinusoidal encoding of the relative offset, to the attention logits, rather than the full four-term decomposition of \citet{dai-etal-2019-transformer}. The models are trained for 15000 steps without warmup and with dropout rate at 0.1. See our repository for more training details.

\subsubsection{Results}

\begin{figure*}
    \includegraphics[width=1\columnwidth]{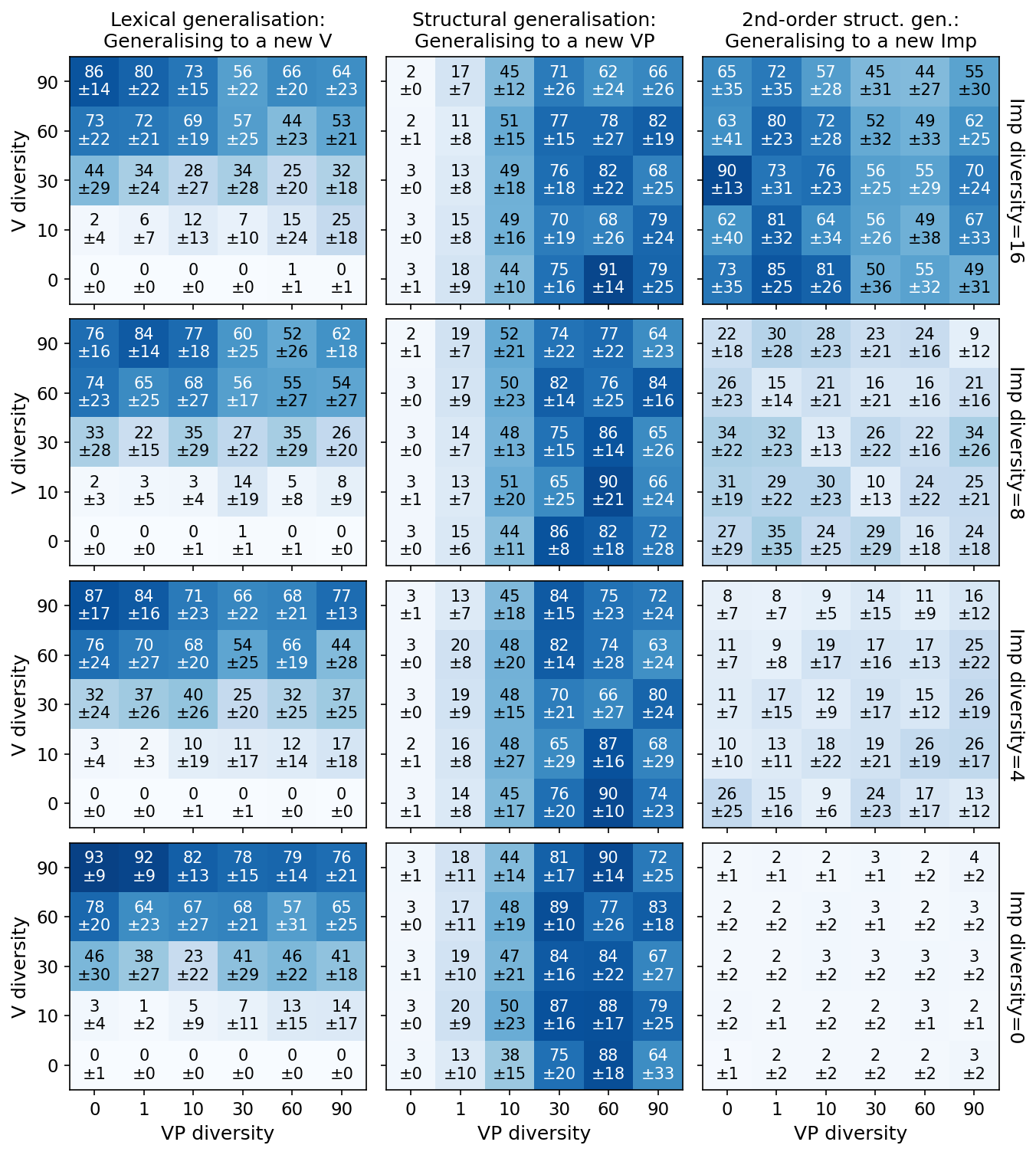}
    \caption{Accuracies (\% $\pm$std) in the SCAN variants per type diversities of the three hierarchical types (Imp(VP(V))).}
    \label{fig:heatmap}
\end{figure*}

Figure~\ref{fig:heatmap}
shows the generalisation accuracy for each of the three types
in different data splits that have an increasing number of other constructors for that type.
Firstly, we see again that the number of constructors of each type correlates strongly with the ability to generalise to a new constructor of that type. We see that performance is poor (around 0-3\%) without much TD and increases somewhere between 60-90\% percent accuracy when the type diversity is increased, in all three generalisation cases.

Importantly, there is no significant difference in how TD and accuracy correlate in the lexical generalisation task and how they correlate in the two structural generalisation tasks. This result is in contradiction with previous works that have concluded that structural generalisation is harder than lexical for Transformers \citep{kim-linzen-2020-cogs,yao-koller-2022-structural,li-etal-2023-slog,jabbar-etal-2025-distinguishing}. The difference to the experiments in these previous works is that in our experiment there is roughly an equal level of type diversity in the lexical and structural types, whereas in the previous works the TD of the structural types is kept significantly
lower than that of the lexical types (see the previous section).


The relationship between accuracy and sub/supertree type diversity is more complicated. For a large part there is no correlation that is more significant than the noise from the variation across random seeds. However, there are some parts of the heatmap where we can see a signal that is not buried under the noise. Generalising to the novel lexical constructor `jump' seems to be aided by VP diversity when V diversity is at 10 constructors, i.e. low but at a level where the models are not failing completely any more, as they are at 0 added constructors. For example, at V diversity=10 and Imp diversity=0, the mean lexical generalisation accuracy is increased from 3\% to 14\% when VP diversity goes from 0 additional constructors to 90. The correlation is directionally similar at other Imp diversity levels. Across the four Imp diversities at V diversity=10, Spearman's $\rho$ between lexical generalisation and VP diversity is 0.86.

Interestingly, when V diversity is higher, the correlation of V generalisation and VP diversity \emph{reverses}: accuracy is higher at \emph{lower} VP diversity values. For example, at V diversity=90 and Imp diversity=0, mean lexical generalisation accuracy decreases from 93\% to 76\% as VP diversity is increased from 0 to 90 new constructors.  Across the four Imp diversities at V diversity=90, Spearman's $\rho$ between lexical generalisation and VP diversity is -0.67.
It seems therefore that when the diversity of V is low and the models are struggling to generalise to `jump', they can be helped by VP diversity too, but when there is plenty of V diversity, VP diversity is only obstructing lexical generalisation: at fixed dataset size, more constructors means fewer samples per each constructor.

The impact of Imp diversity on lexical generalisation is similar to VP diversity, though the positive effect is smaller: Spearman's $\rho$ between lexical generalisation and Imp diversity is 0.28 at V diversity = 10, and -0.75 at V diversity = 90.

Generalising to a new VP is not swayed by the V or Imp diversities; only the VP diversity seems to drive generalising to a new VP in our sweep of training set variants. The correlation of VP diversity and VP generalisation seems to plateau or even reverse after 60 added VP constructors. This may be because, unlike V or Imp constructors, VP constructors can be quite different from one another since their output lengths vary from 2 to 8 tokens. The test case, `opposite' is mapped to three output tokens (\texttt{Adv Adv Verb}).
There are in principle $2^3=8$ constructors of length 3, but we exclude constructors that don't include at least one \texttt{Adv} and one \texttt{Verb}.
In the current experiment, all five of the possible new constructors of length three appear in the variant with 30 new constructors: the variants with 60 or 90 new constructors can only add new VP constructors with longer outputs. This is due to our weighted random sampling of new constructors, in which we weight shorter constructors more to keep the number of short constructors roughly balanced with the number of the long constructors that are more numerous and would dominate the training data if the sampling was uniform across all possible constructors. The dataset variants have the following number of constructors of length four, skewed toward variants with 60 or fewer new constructors: $1/1, \ 2/10, \  5/30, \ 11/60, \ 13/90$. The decreasing relative frequency of constructors of lengths of 3 or 4 might explain the plateauing of the accuracy at 60 new constructors.

In the second-order structural generalisation case, all three type diversities have some impact. Besides the expected correlation of Imp diversity with Imp generalisation, VP diversity correlates with Imp generalisation at Imp diversity=4 (Spearman's $\rho$ = 0.58). V diversity also has a small positive effect at Imp diversity = 0 (Spearman's $\rho$ = 0.49), though this is close to the noise level and could be confounded by the smaller base dataset (see above). At higher Imp diversity levels, the effect turns negative again. At Imp diversity=16, Spearman's $\rho$ for V diversity is -0.44 and for VP diversity is -0.65.
This means, as with \emph{super}tree type diversity for lexical generalisation, \emph{sub}tree type diversity for second-order structural generalisation has a positive effect when the second-order structural diversity itself is low and negative when it is high.

\subsection{Type diversity and the atom/compound divergences of \citeauthor{keysers2019measuring}} \label{sec:expts-dbca}

In the previous sections we have established that type diversity enables compositional generalisation. In Section~\ref{sec:compositionality} we mentioned that there are multiple ways to define what `compositionality' means in `compositional generalisation'. The traditional definition comes from Montague, but \citet{keysers2019measuring} offer an alternative definition: for them, `compositionality' means how much the distributions of combinations of atomic units diverge between the training and test sets.

\begin{figure*}[!h]
\centering
    \includegraphics[width=0.87\columnwidth]{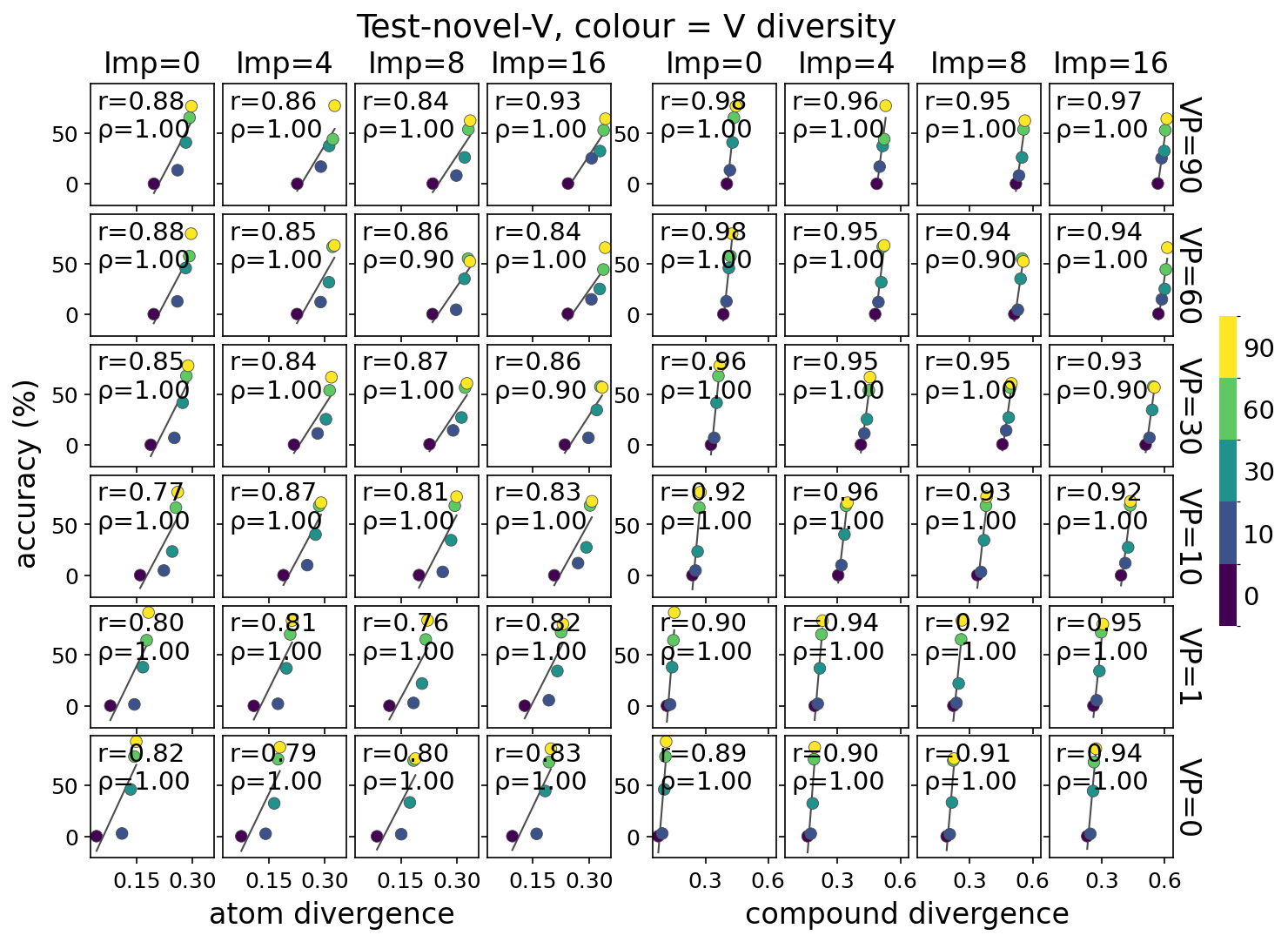}
    \caption{Transformer accuracy as a function of atom and compound divergences (V diversity colour coded) in lexical generalisation (a new V constructor in test set). Over the 24 datasets, mean correlation of accuracy with atom divergence: r=0.84, $\rho$=0.99; with compound divergence: r=0.94, $\rho$=0.99.}
    \label{fig:scan-scatter-v}
\end{figure*}
\begin{figure*}[!h]
\centering
    \includegraphics[width=0.84\columnwidth]{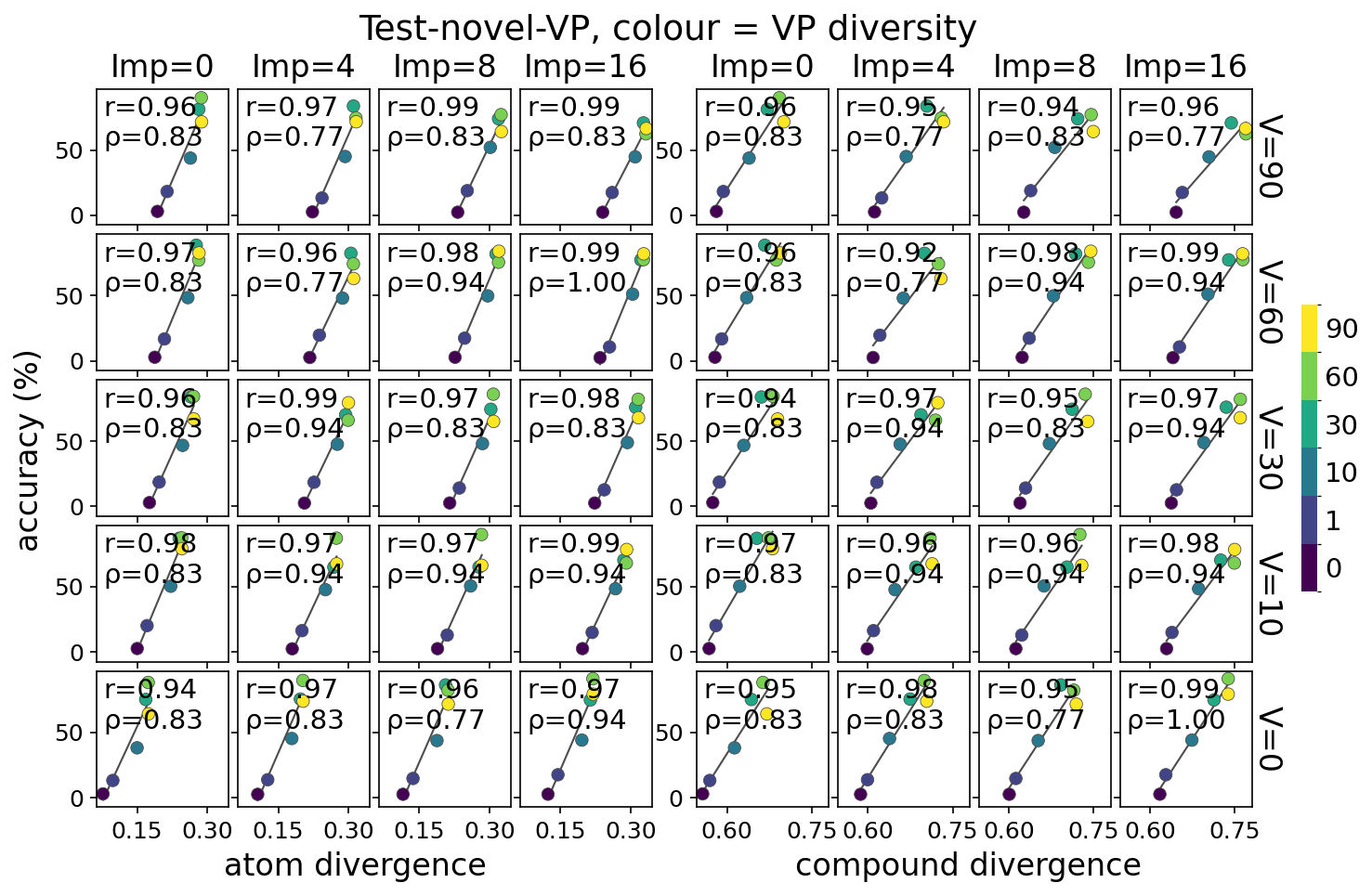}
    \caption{Transformer accuracy as a function of atom and compound divergences (VP diversity colour coded) in structural generalisation (a new VP constructor in test set). Over the 20 datasets, mean correlation of accuracy with atom divergence: r=0.97, $\rho$=0.86; with compound divergence: r=0.96, $\rho$=0.87.}
    \label{fig:scan-scatter-vp}
\end{figure*}
\begin{figure*}[!h]
\centering
    \includegraphics[width=0.92\columnwidth]{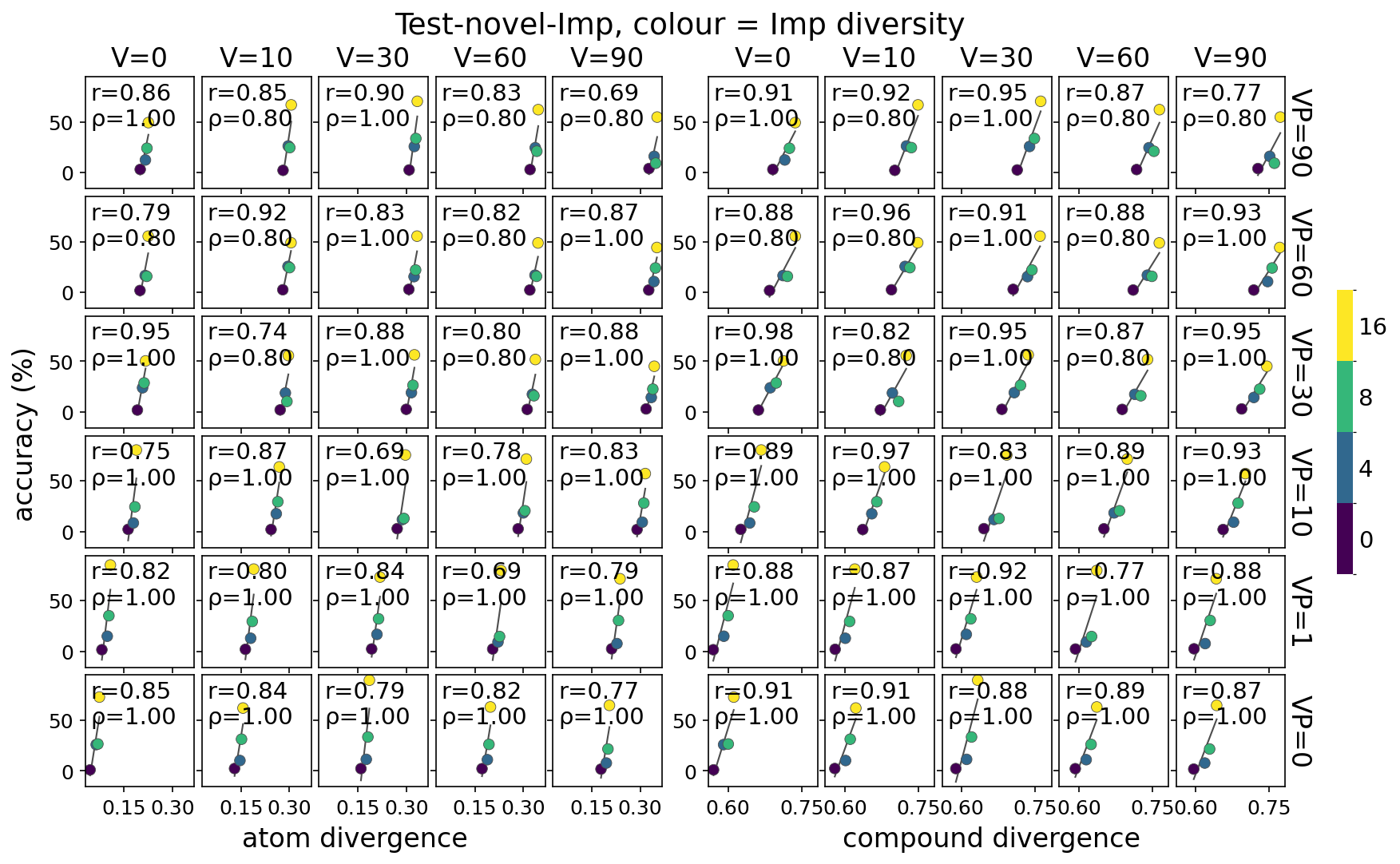}
    \caption{Transformer accuracy as a function of atom and compound divergences (Imp diversity colour coded) in second-order structural generalisation (a new Imp constructor in test set). Over the 30 datasets, mean correlation of accuracy with atom divergence: r=0.82, $\rho$=0.95; with compound divergence: r=0.89, $\rho$=0.95.}
    \label{fig:scan-scatter-imp}
\end{figure*}

Since both the atom/compound divergence of \citet{keysers2019measuring} and our type diversity are properties of datasets, a natural question to ask is, how do the two properties relate to each other? The relationship is straightforward: as increasing type diversity means adding to the training set more unique constructors, this means that both the atom and compound divergence between training and test set increase (see
Appendix~\ref{sec:divergences} for details). 

This is where we run into a contradiction: type diversity correlates with both Transformer performance and the atom and compound divergences, but the main observation of 
\citet{keysers2019measuring} is that, in their CFQ dataset, there is `a strong \emph{negative} correlation between the compound divergence and the mean accuracy' (emphasis ours), which they interpret to mean that `compound divergence seems to capture the core difficulty for these ML architectures to generalize compositionally.'
Our results suggest the opposite: the more compound divergence the better Transformers generalise. Moreover, performance correlates with atom divergence, too, even though it is also assumed to make the task harder in the original work. 
Figures~\ref{fig:scan-scatter-v},~\ref{fig:scan-scatter-vp}, and~\ref{fig:scan-scatter-imp} show the generalisation performance
as functions of atom and compound divergences, in each of the SCAN variants of the previous section: accuracy and the divergences correlate in all variants (Pearson's r and Spearman's $\rho$ reported).
Of course,
the compound divergence in our experiments is of a specific kind, caused 
by the increased diversity of the type of the test case constructor.
Nevertheless, our result is in conflict with the most general hypothesis of \citet{keysers2019measuring} that compound divergence captures \emph{the core} difficulty of compositional generalisation.

In the experiments of \citet{keysers2019measuring}, more compound divergence mostly means that the models need to generalise more: the test set includes more combinations of atoms that don't appear in the training set. In our experiments, the test set is always the same and compound divergence is increased by adding diversity into the training set instead of making the test case more dissimilar to the training samples: the test case still needs `as much' generalisation, but the training set provides a broader support for it.
The fact that more compound divergence can mean `more generalisation is needed to process the test cases' in one instance, and `training set provides a broader support for the test cases' in another instance suggests that the notion of \emph{compound divergence} is too vague to be used as a general method to assess the compositionality of training-test set splits, and that the difficulty of compositional generalisation is explained by compound divergence only in some specific instances.

\section{Experiments and results on the SLOG dataset} \label{sec:expts-slog}

In this section we return to assessing generalisation in the semantic parsing task.
Instead of COGS, which only has one proper structural generalisation case (cf. Footnote~\ref{fn:cogs-struct}), here we use the SLOG dataset \citep{li-etal-2023-slog}, which has the same task format but focuses on structural generalisation.
Section~\ref{sec:expts-cogs} showed how adding type diversity can increase Transformer accuracy in the structural generalisation task of COGS from around 2\% to over 90\%. However, we noticed that besides type diversity, a property of abstract syntax, also some concrete syntax properties affect the Transformer performance. As discussed in Section~\ref{sec:surface},
WMP
showed that such surface properties as variable names and the orders of the tokens in input and output sequences can have a surprisingly large effect on Transformer performance. In the same vein, in this section we assess the effect of changes to the semantic format and input tokenisation. Since our main thesis is about type diversity, another motivation for these experiments is to inspect whether the changes in surface properties confound the results regarding type diversity in the previous sections.

In Section~\ref{sec:expts-scan} we showed in more detail, although in the simpler task of SCAN, how sub- and supertree type diversities help generalisation (only) when type diversity itself is low.
In the GF grammar of natural language, there are types that have only one or two constructors, which means there cannot be much type diversity even in principle. In these cases the  sub- and supertree type diversity becomes especially important.
We previously noted that \texttt{CN, Det,} and \texttt{Num} diversities collapse into \texttt{NP} diversity since they do not appear anywhere outside NPs; this is the extreme case of when subtree type diversity is important. 

In this section we explore how these factors, among others, play out in the suite of structural generalisation tasks of the SLOG benchmark.


\subsection{The SLOG dataset}

SLOG includes test cases with varied generalisation targets.
We divide the test cases of SLOG into three categories, based on the type (NP, VP, or clause) of the constructor in a novel
position.\footnote{In addition to the ones mentioned, there are four other test cases in SLOG that we exclude: three deeper-nesting-depth cases have the same out-of-vocabulary problem as COGS (see Footnote~\ref{fn:cogs-struct}), and in the `Active voice wh-questions' cases Transformers achieve always close to 100\% accuracy, making it less interesting than the harder cases.}
We use acronyms for noun phrase NP, common noun CN, verb phrase VP, prepositional phrase PP, relative clause RC, participial adjectival phrase PAP, complement clause CC.
In this section, we call CN constructors `NP constructors' for simplicity; see discussion in Sections~\ref{sec:type-diversity} and~\ref{sec:expts-cogs}.
\newline
\textbf{(1) NP constructor in a new role}.
`PP/RC in subject/indirect object NPs',
`Shallower PP recursion / center embedding' and
`PP/RC in wh-questions' are all similar in the sense that in all seven cases an NP with a PP or RC is in a new position: as subject, as indirect object, in a 3rd-order nested PP/RC, or in a question sentence.
\newline
\textbf{(2) Slashed VP constructor in a new role.}
The `Indirect object-extracted RC' sentences have relative clauses with ditransitive verbs, and the entity linked to the main clause is in the \emph{indirect} object, e.g. `Noah saw the cat that Emma gave a cake to \_', leaving a gap, denoted `\_' in the example. Training set includes relative clauses with 
slashed\footnote{`Slashed' as in e.g. VP/NP in combinatory categorial grammar, meaning the verb phrase is missing a noun phrase: for example `liked' is a transitive VP missing the object. Technically the VP types  are almost always slashed in GF, and in gapless sentences the gap is filled by the parent constructor; here, by  `slashed' we mean structures where the gap is not filled by the parent constructor and the gap remains in the complete sentence.} ditransitive VPs with the gap in the \emph{direct} object slot: `Benjamin snapped a cookie that Michael served \_ to Olivia.'.
Similarly, in the `3-place direct/indirect object wh-questions', the questioned role is the direct/indirect object of a 3-place verb (`What did Emma give \_ to the cat?'). The training set includes only questions with transitive verbs, and only in the active voice, and the questioned role is either the subject (`Who ate the cake?') or the direct object (`What did Emma eat?').
The `Passive voice wh-questions', can be a transitive verb like in the training set, 1-place unaccusative or unergative verb, or a 3-place verb, and they are in the passive voice which never appears in questions in training.
\newline
\textbf{(3) (Slashed) Clause constructor in a new role.}
The `Shallower Tail complement clause (CC)
recursion'\footnote{It is called `complement phrase (CP)' in the SLOG paper, but CC is more accurate name since it is a complete clause, e.g. `the cat slept' in a sentence `Emma saw that the cat slept'.}
is similar to the nested PPs and RCs, but also different in the sense that it doesn't introduce an NP constructor but a complement clause
in a new role.
In the `Complement clause (CC) wh-questions' case,
a slashed complement clause is in the verb phrase of a question sentence, and
the gap is the object NP (`Whom did a governor hope that William froze?').

Here we should flag some limitations of looking only at type diversity. There are symmetries in the data beyond types, which models can exploit to generalise compositionally. This also depends on the grammar (cf. Section~\ref{sec:type-diversity}): whether VP and slashed VP are separate types, as they are in GF unlike in CCG.
Even though there are no constructors for slashed complement sentences in the training set, the learner might see how other types, mainly VPs, are slashed, and infer that the complement sentence could be slashed too. That is, the learner might generalise \emph{across} types, besides generalising \emph{within} types from one constructor to another, in case some of the symmetries of language (e.g. between VP and clause and slashed VP and slashed clause) are not captured by the set of types in the grammar. 

We should also flag some of the complexity of the patterns present even in this small quasi-natural language fragment.
There is variation and ambiguity in SLOG that obscures systematicity (although not nearly as much as in real natural language), making learning the symmetries harder. Perhaps the most conspicuous piece of variation is the double-object-construction (DOC) / to-construction alternation of 3-place verbs (`give a dog a bone' / `give a bone to a dog'). Word-level ambiguity includes using `that' both as a relative pronoun and a complementiser; using both the causative and the inchoative meanings of verbs like `break', which map the subject to different thematic role predicates (Agent vs Theme); and using `to' both as a preposition (`give a bone to a dog') and as an infinitive marker (`a dog wanted to bark').
Although SLOG was meant to be unambiguous at the sentence level, there are also a few sentences that have a possible alternative parsing and interpretation besides the one in the dataset.
Fortunately, there are not many of them so they do not make a big difference to the results; see Appendix~\ref{sec:ambigous-slog}.

\subsection{Results}

\begin{figure}
\centering
\begin{subfigure}{\textwidth}
    \includegraphics[width=1.00\textwidth]{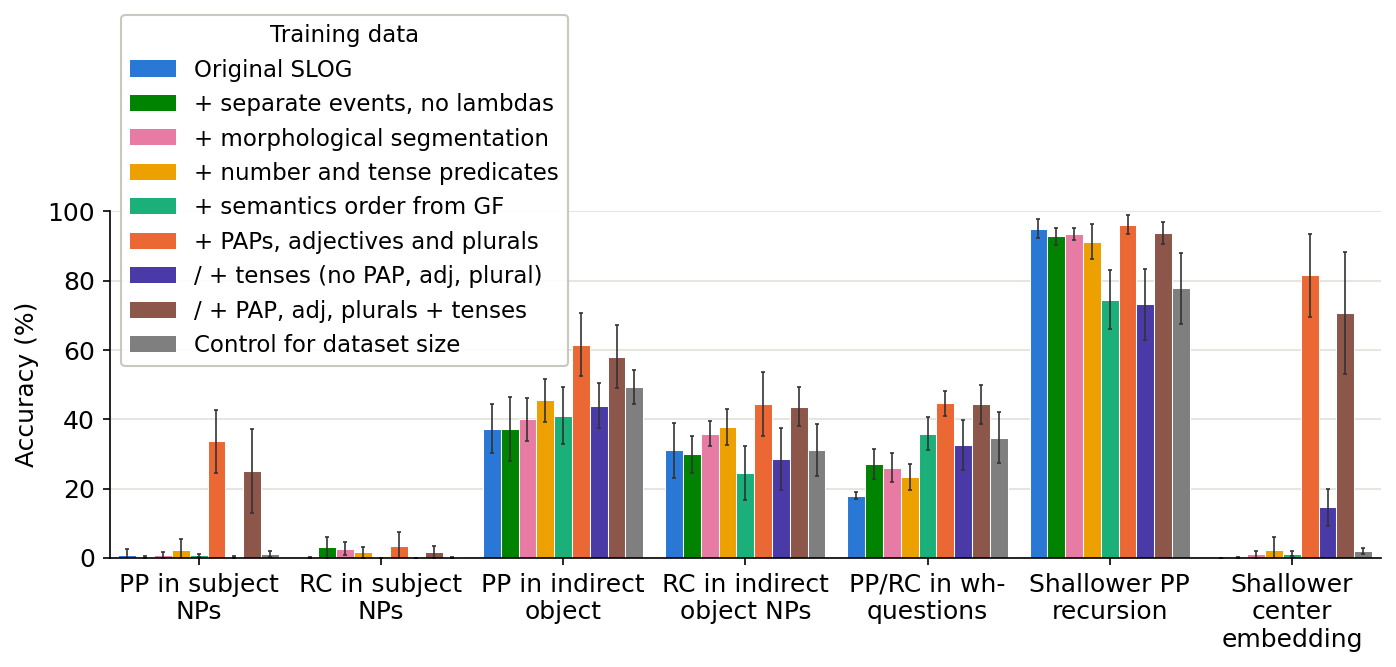}
\end{subfigure}
\begin{subfigure}{\textwidth}
    \includegraphics[width=1.00\textwidth]{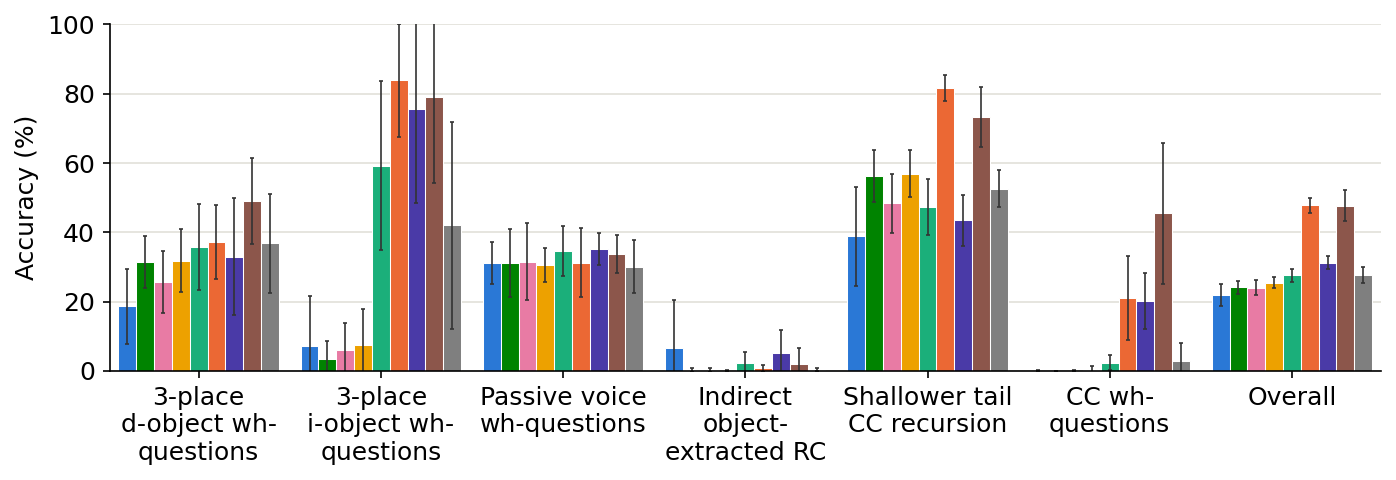}
\end{subfigure}
\caption{Transformer accuracies, trained on different training set variants of SLOG.} \label{fig:slog-results}
\end{figure}

Figure~\ref{fig:slog-results} shows the SLOG accuracies for Transformers trained on different  training sets.
The added diversity increases the dataset size, so we include an augmented dataset, similar to that in Section~\ref{sec:expts-cogs}, and report it as `Control for dataset size' in the figure.



\subsubsection{Separating events, renaming variables, removing explicit lambdas}

In the first group of changes, we simplify the logical semantics format.
The event predicates are separated from the thematic role predicates, and the noun modifier \texttt{nmod} predicates do not include the noun. The variables get their names from the GF syntax for higher-order bound variables, e.g. \texttt{v0}. For example, the semantic of `Who rented a sandwich in a house to Emma?' in the original SLOG is:

\begin{small} \begin{verbatim}
rent . agent ( x _ 1 , ? ) AND rent . theme ( x _ 1 , x _ 3 ) AND
    rent . recipient ( x _ 1 , Emma ) AND sandwich ( x _ 3 ) AND
    sandwich . nmod . in ( x _ 3 , x _ 6 ) AND house ( x _ 6 )
\end{verbatim} \end{small}
which is changed to
\begin{small} \begin{verbatim}
rent ( v0 ) AND Agent ( v0 , ? ) AND Theme ( v0 , v1 ) AND
    Recipient ( v0 , Emma ) AND sandwich ( v1 ) AND
    Nmod . in ( v1 , v2 ) AND house ( v2 )
\end{verbatim} \end{small}
We also remove the explicit lambda functions from the primitives, since they don't appear in the semantics of whole sentences either. An original SLOG semantics for the primitive verb `admire':
\begin{small} \begin{verbatim}
LAMBDA a . LAMBDA b . LAMBDA e . admire . agent ( e , b ) AND
    admire . theme ( e , a )
\end{verbatim} \end{small}
is changed to
\begin{small} \begin{verbatim}
admire ( v0 ) AND Agent ( v0 , v1 ) AND Theme ( v0 , v2 )
\end{verbatim} \end{small}

These changes mostly have little effect on the results, but there are a couple of cases where they do. One affected case is
3-place direct object wh-questions,
where a systematic error is to omit either the theme or the recipient, e.g.
`What did Liam lend to Lucas ?' \ \ $\rightsquigarrow$ \ \ \texttt{lend . agent ( x \_ 3 , Liam ) AND lend . theme ( x \_ 3 , ? )}, missing the recipient \texttt{AND lend . recipient ( x \_ 3 , Lucas )}.

In the new format \ \ \ \ \texttt{lend ( v0 ) AND Agent ( v0 , Liam ) AND Theme ( v0 , ? ) AND Recipient ( v0 , Lucas )} especially the separated event and thematic predicates could make the task easier for the models. When the verb is not included in each of the thematic role predicates the interpretation is more \emph{compositional}. For example, if the verb has to be included in the agent predicate, we cannot have just a single definition of the \texttt{iVP} function for passive voice verb phrases that calls an \texttt{iAdvVP} function, also with a single definition, as defined Appendix~\ref{sec:data-generation-cogs-slog}:

\begin{small} \begin{verbatim}
fun iVP : VP -> Ind -> Event -> Prop ;
def iVP (AdvVP vp adv) = iAdvVP adv (iVP vp) ;

fun iAdvVP : Adv -> (Ind -> Event -> Prop) -> Ind -> Event -> Prop ;
def iAdvVP (PrepNP by_Prep np) vpf = \i ->
        iNP np (\y,e -> And (vpf i e) (Agent y e)) ;
\end{verbatim} \end{small}
Instead, we would need to do pattern matching on the \texttt{vp} argument in the definition \texttt{def iVP (AdvVP vp adv)} so that we can pass the verb from inside the \texttt{vp} to the \texttt{Agent} function, adding many more definitions. Presumably, a less compositional interpretation such as this makes compositional generalisation harder for seq2seq models.

\subsubsection{Morphological segmentation}

For example `painted' is segmented to `paint ed'.
This modification makes little difference to the results if there are only past tenses in the data, but a large difference after more verb tenses are added (see below). 

\subsubsection{Adding number and tense predicates} Before adding plural forms and new tenses, we add predicates for tense and number that are always number=singular, time=past, anteriority=simultaneous, e.g.:

\begin{small} \begin{verbatim}
rent ( v0 ) AND Agent ( v0 , ? ) AND Theme ( v0 , v1 ) AND
    Recipient ( v0 , Emma ) AND Anteriority ( v0 , Simul ) AND
    Time ( v0 , Past ) AND sandwich ( v1 ) AND Nmod . in ( v1 , v2 )
    AND Number ( v1 , Sg ) AND house ( v2 ) AND Number ( v2 , Sg )
\end{verbatim} \end{small}
Like segmentation, this has little impact on the results, but is a prerequisite for adding new tenses and plural forms. The fact that these modifications make almost no difference to the performance means that they do not confound the results of adding the new tenses and plural forms to the dataset, which is the step that increases type diversity.

\subsubsection{Changing the order of semantic terms}

WMP
noticed that the order of the terms has an effect especially in the PP-in-Obj$\rightarrow$PP-in-Subj task of COGS, but they always changed also the natural language sentence, and always had the order of semantics aligned with the sentence. In contrast, here we test changing only the output sequence order, leaving input sequence unchanged.

The interpretation by computation described in
Appendix~\ref{sec:data-generation-cogs-slog}
creates an abstract syntax tree of the semantics, which we flatten in the linearisation to resemble the COGS/SLOG format. One consequence of this semantic parsing method is that the order of the semantic terms does not necessarily align with the order tokens in the natural language. Since the semantics is a flat conjunction, any order is logically equivalent within the sets of assertions and presuppositions. For the previous setups, we have reordered the terms as a post-processing step, so that the order is the same as in COGS/SLOG, i.e. aligned with the word order in the sentence. Here we leave the order of the terms as they are in the linearised GF abstract syntax.
This order is guaranteed to be compositional since it is the output of interpretation and linearisation in GF \citep{ranta2004computational}.

The case where the order has the largest effect is the indirect object wh-questions case. Here, the systematic error for the models trained and tested on original, aligned order is that the interrogative variable is placed in the \texttt{Theme} predicate when it should be in the \texttt{Recipient}. For example, the semantics of `Whom did a boy return a jacket to?' are for the aligned order and the GF order, respectively:

\begin{small} \begin{verbatim}
boy ( v1 ) AND Number ( v1 , Sg ) AND return ( v0 ) AND
    Agent ( v0 , v1 ) AND Theme ( v0 , v2 ) AND Recipient ( v0 , ? )
    AND  Anteriority ( v0 , Simul ) AND Time ( v0 , Past ) AND
    jacket ( v2 ) AND Number ( v2 , Sg )

boy ( v1 ) AND Number ( v1 , Sg ) AND jacket ( v2 )
    AND Number ( v2 , Sg ) AND return ( v0 ) AND Agent ( v0 , v1 ) AND
    Theme ( v0 , v2 ) AND Recipient ( v0 , ? ) AND
    Anteriority ( v0 , Simul ) AND Time ( v0 , Past )
\end{verbatim} \end{small}
The typical mistake for the aligned-order models is to output, incorrectly `\texttt{Theme ( v0 , ? )}' and then right after that output, correctly `\texttt{Recipient ( v0 , ? )}': the question mark is outputted more than once (which is incorrect) in 63.2\% samples by aligned-order models, and only 18.1\% by GF-order models. It seems two things are important here. Firstly, `\texttt{?}' is never in the Recipient in the training set, which is of course the case regardless of the order of the terms. What makes the aligned-order test cases harder, it seems, is that the \texttt{Theme} and \texttt{Recipient} predicates are outputted before the noun predicate (here `\texttt{jacket ( v2 )}'). When the aligned-order model gets to the \texttt{Theme} predicate, it would need to output `\texttt{?}' even though in every wh-question in training a role slot whose entity had not yet been introduced was `\texttt{?}'. In the compositional GF-order, the semantics of the direct object (jacket) have already been outputted when the theme and recipient role predicates need to be filled with entities, making it easier to fill the roles correctly.

\subsubsection{NP diversity}
    %


Similarly to Section~\ref{sec:expts-cogs}, the NP type is diversified with NPs that include adjectives and/or participial adjectival phrases (PAPs), and the grammatical number of half of the common nouns is switched to plural from the original singular. The three additions are independent of each other: an NP can include 0, 1, 2, or 3 of these edits. For example, if the NP `a cat' gets all three additions, it could become `happy cats sitting on a table'. 

As in the COGS experiments, the `PP in subject NPs' accuracy improves, from nearly zero to about 35\%. Results in the six other test cases belonging to the group (1)
are improved too, although the magnitude of the change varies a lot. Shallower centre embedding has the largest improvement, from near zero to around 80\%. `RC in subject NPs' has the smallest improvement: from 0.0 (+-0.0) to 1.8 (+-1.7)\%, but increasing the accuracy from exactly zero to a non-zero value is a significant improvement. However, adding the PAPs, adjectives, and plural forms is not the only way to improve the results in this test case: the accuracy is non-zero also in the three setups where the order of the semantic terms is the original, and is reduced back to zero when we change it to the GF-generated order. 


\subsubsection{Clause diversity}



Similarly to Section~\ref{sec:expts-cogs}, we add seven new verb tenses to the dataset.
New tenses add subtree type diversity for clauses (or sentences).
There are two test cases that are of type \emph{clause} in SLOG: the \emph{CC wh-questions} and \emph{Shallower tail CC recursion}. Adding tenses improves the result in the former but not in the latter.
Furthermore, in the \emph{CC wh-questions} case, the combination of the NP diversity (PAPs, adjectives, and plurals) and the new tenses has the largest positive effect on the results.
The main property of this test case is a gap that is far away from the wh-word that fills it: `\emph{What} did Emma say that the cat  found \_?', which is why the case is called `Wh-questions long movement' in the original paper.  This seems to make variable binding difficult. A difference in the predictions comparing the dataset with only PAPs, adjectives, and plurals and the one with also the tenses is that without tenses binding the correct variable to the anteriority and time predicates is less accurate. For example, the sentence `What did Liam notice that Benjamin ate?' has the semantics:

\begin{small}
\begin{verbatim}
notice ( v0 ) AND Agent ( v0 , Liam ) AND Ccomp ( v0 , v1 ) AND
    eat ( v1 ) AND Agent ( v1 , Benjamin ) AND Theme ( v1 , ? )
    AND Anteriority ( v1 , Simul ) AND Time ( v1 , Past ) 
    AND Anteriority ( v0 , Simul ) AND Time ( v0 , Past ) 
\end{verbatim} \end{small}
A common mistake for the models trained without the additional tenses is to bind the wrong variable to the time and anteriority predicates that should have `\texttt{v1}'. If the training set only contains Anteriority=Simul, Time=Past, variable binding for these predicates has not been crucial, not encouraging to learn to bind the variables carefully. For the best performing random seeds, the setup without tenses has various kinds of variable binding errors for 25.0\% of all errors in this test case, whereas the setup with tenses has 0.0\%, meaning that variable binding is never the sole mistake though it sometimes appears together with other mistakes. Pooled over all seeds the respective numbers are 22.9\% and 2.6\%.

One notable detail is that when tense is changed from the original past-simultaneous to, for example, present-simultaneous, the subject-predicate agreement becomes relevant: `a mouse walked' and `mice walked' have the same verb form, but `a mouse walks' and `mice walk' do not.
This probably makes it easier to learn to assign the correct variable to the time and anteriority predicates when the training set includes both plural forms \emph{and} all tenses.

\subsection{Conclusions from the SLOG experiments}

In this section, we explored the effects of various data properties; in addition to type diversity, properties such as output sequence order and separating event predicates and thematic role predicates have a large impact on the results in some cases.
In total, mostly due to increased NP diversity, the average accuracy was increased from 27.7\% to 47.7\%.

We emphasise the complexity of the natural language data properties beyond type diversity.
Although type diversity is still a major factor affecting the results,
natural language has an intricate structure that is not explained by simplified metrics alone.
For example, the subject-verb agreement becomes relevant only when both the plural forms of nouns and different tenses of verbs are added to SLOG: this is not captured by looking at the diversities of the two types in isolation. 
This also highlights the risks of using simplified diagnostic datasets that do not include this intricate structure and diversity of natural language; conclusions drawn from these datasets do not necessarily generalise to models trained on real, diverse natural language datasets.

\section{Discussion on compositional generalisation benchmarking methods} \label{sec:discussion}



How should compositional behaviour in Transformers be assessed?
When the dataset is natural language, it can be assumed to have compositional structure.
However, natural languages have a number of other properties, too, such as synonymy, polysemy, idioms, irregular verb inflections, constructions etc.
Because of the black-box nature of deep neural networks, it can be difficult to tell whether a neural NLP model has suboptimal performance because it has not learned the systematic, compositional structure, or because it has not learned all the other properties of the natural language that can obscure systematicity.
Therefore, to assess compositional generalisation, these other properties need to be controlled for.

A popular approach to control for the other properties has been \emph{diagnostic datasets}, such as the SCAN and COGS datasets used in this work, that exclude most of the other properties from the data, focusing on compositionality.
These are influential benchmarks that have been widely used to assess the compositional generalisation capacity of seq2seq models.
However, multiple studies have since questioned the robustness of the results and conclusions drawn from the diagnostic datasets, of which we have already described some.
\citet{bastings-etal-2018-jump} pointed out that, unlike in natural language,
there are few target-side dependencies in SCAN, which allows simple models to perform well without using composition in any interesting way, and that `their performance is therefore not a realistic indicator of their generalization capability'.
\citet{patel-etal-2022-revisiting} added more verbs to SCAN and found that Transformers achieve close to perfect accuracy on the `add jump' split, showing that the original results were misleading.
WMP
showed that incidental properties of the logical form in COGS affect the results significantly.
In light of these results, it is not surprising that a group of benchmarks
that all aim to assess compositional generalisation have been shown to rank models in different order \citep{sun-etal-2023-validity}, which betrays a poor \emph{construct validity} of compositional generalisation in (at least some of) these benchmarks.
Our own results suggest that the COGS dataset has been misleading with regards to Transformers' capacity for structural generalisation because of its unnaturally low type diversity for the tested structure (NP).

These results point to the limitations of trying to simply remove all factors other than systematic compositionality from a dataset. It is not possible, since \emph{systematic compositionality is an abstract property that has to be instantiated in an actual dataset, and therefore co-exists necessarily with other properties of the dataset}. 
Instead of removing other factors from a dataset, we need to map the variables that affect compositional generalisation, and control for their effect when comparing how different models perform.
%
%
In this work we controlled for a few factors, mainly the specific kind of diversity we call type diversity, but mapping the effects on generalisation of real natural language properties such as variation and ambiguity remains an interesting topic for future work.

Importantly, what our results do \emph{not} show is that Transformers would reach perfect or human-level compositional generalisation ability given enough type diversity. Accuracy typically plateaus before reaching exactly 100\%, in line with previous works that have added lexical diversity and assessed lexical generalisation \citep{patel-etal-2022-revisiting,zhou-etal-2023-data}. 
This is congruent also with works that have assessed neural network capacity to learn grammar, and found that they typically learn many aspects of grammar but do not learn to use it \emph{systematically}, i.e. to categorically delineate grammatical from ungrammatical sequences based on the rules that humans adhere to \citep{linzen-etal-2016-assessing,lan2024large,moisio-etal-2024-llms}.

\section{Summary and Conclusions}

In this work we:
    (1) Generated linguistically diverse and in this sense more realistic variants of previously published COGS and SLOG datasets. The data generation method, based on Grammatical Framework, is extendable to still more linguistic structures and different languages in possible future work.
    (2) Showed that, contrary to conclusions in some previous work, structural generalisation is not significantly harder than lexical for Transformers, but that this previous result is explained by 
    lower type diversity of the tested structural types compared to the lexical types.
    (3) Found that sub- and supertree type diversity can also help generalisation when type diversity itself is low.
    (4) Showed that compound and atom divergences can correlate with Transformer generalisation performance (contrary to \citet{keysers2019measuring}), if divergence is caused by increased type diversity.
    (5) Demonstrated effects of other data properties such as the order of the semantic terms on generalisation results.

\section{Limitations}

We used only Transformer models, since this has become the dominant architecture in NLP.
Our results probably would not generalise to all seq2seq models, since prior work has reported quite large differences between Transformers and, for example, LSTM networks \citep{kim-linzen-2020-cogs}. Moreover, we used only one hyperparameter setup per experiment.
The results would undoubtedly change at least a little if the hyperparameters were changed.
This leaves the possibility, in principle, that some of the conclusions are specific to the Transformer hyperparameters that we used. However, we took the hyperparameters
from previous work to make the results comparable.
Relatedly, the variance across random seeds is very high in most of our results. We have reported the standard deviation of the variance, and have attempted to base our conclusions only to results that are apparent notwithstanding the high variance.


\appendix

\clearpage
\section{Generating SCAN} \label{sec:data-generation-scan}

In GF, the SCAN grammar can be defined by the abstract syntax in Table~\ref{tab:scan-abstract}.
The concrete syntaxes of the target and source language are shown in Tables~\ref{tab:scan-input}~and~\ref{tab:scan-output}.
\begin{table}[h]
\begin{small} \begin{verbatim}
abstract Scan = {
    flags startcat = C ;
    cat
        C ; S ; Imp ; VP ; Verb ; Adv ; VD ; V ;
    fun
        UseConjImp  : S             -> C ;
        UseImp      : Imp           -> C ;
        And         : Imp -> Imp    -> S ;
        After       : Imp -> Imp    -> S ;
        ImpVP       : VP            -> Imp ;
        Twice       : VP            -> Imp ;
        Thrice      : VP            -> Imp ;
        UseV        : V             -> VP ;
        DirVP       : Verb -> Adv   -> VP ;
        OppositeVP  : Verb -> Adv   -> VP ;
        AroundVP    : Verb -> Adv   -> VP ;
        VVerb       : V             -> Verb ;
        VDVerb      : VD            -> Verb ;
        left_Adv, right_Adv             : Adv ;
        turn_VD                         : VD ;
        walk_V, run_V, jump_V, look_V   : V ;
}
\end{verbatim} \end{small}
\caption{SCAN abstract syntax.}\label{tab:scan-abstract}\end{table}

\begin{table}[h]
\begin{small} \begin{verbatim}
concrete ScanInput of Scan = {
    lincat
        C, S, Imp, VP, Verb, Adv, VD, V = Str ;
    lin
        UseConjImp  i       = i ;
        UseImp      i       = i ;
        CoordImp    i1 i2   = i1 ++ "and" ++ i2 ;
        CoordImpInv i1 i2   = i1 ++ "after" ++ i2 ;
        ImpVP       vp      = vp ;
        Twice       vp      = vp ++ "twice" ;
        Thrice      vp      = vp ++ "thrice" ;
        UseV        v       = v ;
        DirVP       v adv   = v ++ adv ;
        OppositeVP  v adv   = v ++ "opposite" ++ adv ;
        AroundVP    v adv   = v ++ "around" ++ adv ;
        VVerb       v       = v ;
        VDVerb      v       = v ;
        left_Adv            = "left" ;
        right_Adv           = "right" ;
        turn_VD             = "turn" ;
        walk_V              = "walk" ;
        run_V               = "run" ;
        jump_V              = "jump" ;
        look_V              = "look" ;
}
\end{verbatim} \end{small}
\caption{SCAN concrete syntax for the source domain.}\label{tab:scan-input}\end{table}

\begin{table}[h]
\begin{small} \begin{verbatim}
concrete ScanOutput of Scan = {
    lincat
        C, S, Imp, VP, Verb, Adv, VD, V = Str ;
    lin
        UseConjImp  i       = i ;
        UseImp      i       = i ;
        CoordImp    i1 i2   = i1 ++ i2 ;
        CoordImpInv i1 i2   = i2 ++ i1 ;
        ImpVP       vp      = vp ;
        Twice       vp      = vp ++ vp ;
        Thrice      vp      = vp ++ vp ++ vp ;
        UseV        v       = v ;
        DirVP       v adv   = adv ++ v ;
        OppositeVP  v adv   = adv ++ adv ++ v ;
        AroundVP    v adv   = adv ++ v ++
                              adv ++ v ++
                              adv ++ v ++
                              adv ++ v ;
        VVerb       v       = v ;
        VDVerb      _       = "" ;
        left_Adv            = "I_TURN_LEFT" ;
        right_Adv           = "I_TURN_RIGHT" ;
        turn_VD             = "" ;
        walk_V              = "I_WALK" ;
        run_V               = "I_RUN" ;
        jump_V              = "I_JUMP" ;
        look_V              = "I_LOOK" ;
}
\end{verbatim} \end{small}
\caption{SCAN concrete syntax for the target domain.}\label{tab:scan-output}\end{table}

\clearpage
\section{Logical semantics} \label{sec:logical-semantics}

Functions can be declared to be \emph{constructors} by replacing \texttt{fun} with \texttt{data}, which enables using them as constructor patterns in \emph{function definitions} \texttt{def}. Function definitions are needed for \emph{computation}, which in our work is used to interpret syntax trees in logical semantics. In the neo-Davidsonian semantics that we use,
the categories are events, individuals (\texttt{Ind}) and propositions (\texttt{Prop}). The neo-Davidsonian thematic role predicates Agent, Theme, and Recipient are propositions about an individual and an event:
\begin{small} \begin{verbatim}
abstract Logic = {
    cat Prop ; Ind ; Event ;
    fun
        Exist                   : (Ind -> Prop) -> Prop ;
        ExistEvent              : (Event -> Prop) -> Prop ;
        And                     : Prop -> Prop -> Prop ;
        Agent, Theme, Recipient : Ind -> Event -> Prop ;
        Unique                  : (Ind -> Prop) -> Ind -> Prop ;
}
\end{verbatim} \end{small}
\texttt{Exist}, \texttt{ExistEvent}, and \texttt{Unique} are higher-order functions: they take a function (of respective types \texttt{Ind -> Prop}, \texttt{Event -> Prop}, and \texttt{Ind -> Prop}) as their argument.
An \emph{interpretation function} \texttt{iX} is implemented for each category (i.e. type) \texttt{X} in the natural language abstract syntax, which guarantees compositionality of interpretation \citep{ranta2004computational}. The function definitions are applied to a given a natural language syntax tree to compute its interpretation in the neo-Davidsonian semantics.
For example, 
interpretation functions for sentence \texttt{S}, determiner \texttt{Det}, noun phrase \texttt{NP}, verb phrase \texttt{VP}, and verb \texttt{V} could be:

\begin{small}
\begin{small} \begin{verbatim}
abstract Semantics = Logic, Lang ** {
    fun iS : S -> Prop ;
    def iS (Sentence np vp) = ExistEvent (iNP np (iVP vp)) ;

    fun iDet : Det -> (Ind -> Prop) ->
                      (Ind -> Event -> Prop) -> Event -> Prop ;
    def
        iDet a_Det   nf vpf = \e -> Exist (\x -> And (nf x) (vpf x e)) ;
        iDet the_Det nf vpf = \e -> Exist (\x -> And 
                                (Unique (\z -> nf x) x) (vpf x e)) ;

    fun iNP : NP -> (Ind -> Event -> Prop) -> Event -> Prop ;
    def iNP (Nounphrase det n) vpf = iDet det (iN n) vpf ;

    fun iVP : VP -> Ind -> Event -> Prop ;
    def
        iVP (Verb2phrase v np) = \i -> iNP np (\y -> iV2 v i y) ;
        iVP (Verb1phrase v)    = iV1 v ;

    fun
        iV2 : V2 -> Ind -> Ind -> Event -> Prop ;
        iV1 : V1 -> Ind        -> Event -> Prop ;
    def
        iV2 (V2Verb v) subj obj e = And (iV v e) (And 
                                        (Agent subj e) (Theme obj e)) ;
        iV1 (V1Verb v) subj     e = And (iV v e) (Agent subj e) ;

    fun iV : V -> Event -> Prop ;
    fun iN : N -> Ind -> Prop ;
}
\end{verbatim} \end{small}
\end{small}
With these functions, the abstract syntax tree of the sentence `a man drinks a coffee' can be interpreted as, i.e. computed into (`==>' denoting computation below), using the GF command \texttt{put\_tree -compute}, an abstract syntax tree of the logical semantics proposition:
\begin{small} \begin{verbatim}
iS (Sentence (Nounphrase the_Det man_N)
             (Verb2phrase (V2Verb drink_V) (Nounphrase a_Det coffee_N)))
==>
ExistEvent (\v0 -> Exist (\v1 -> And
    (Unique (\v2 -> iN man_N v1) v1)
    (Exist (\v2 -> And
        (iN coffee_N v2)
        (And (iV drink_V v0) (And (Agent v1 v0) (Theme v2 v0)))))))
\end{verbatim} \end{small}
This abstract syntax tree could then be linearised, if needed.

\section{Generating COGS and SLOG} \label{sec:data-generation-cogs-slog}
We re-generate COGS and SLOG, using GF, with slightly modified logical form.
Our interpretation functions of sentence S, tense, anteriority Ant, polarity Pol, determiner Det, verb phrase VP, noun phrase NP, and verb V are similar to the following (slightly simplified, for the full semantics see the Github repository):

\begin{small}
\begin{small} \begin{verbatim}
fun iS : S -> Prop ;
def iS (UseCl (TTAnt t ant) p (PredVP np vp)) =
        ExistEvent (iTense t (iAnt ant (iNP np (iPol p (iVP vp))))) ;

fun
    iTense  : Tense -> (Event -> Prop) -> Event -> Prop ;
    iAnt    : Ant   -> (Event -> Prop) -> Event -> Prop ;
    iPol    : Pol -> (Ind -> Event -> Prop) -> Ind -> Event -> Prop ;
def
    iTense t p = \e -> And (p e) (Time t e) ;
    iAnt ant p = \e -> And (p e) (Anteriority ant e) ;
    iPol PPos vpf = vpf ;
    iPol PNeg vpf = \i,e -> Not (vpf i e) ;

fun iDet : Det -> (Ind -> Prop) ->
                  (Ind -> Event -> Prop) -> Event -> Prop ;
def
    iDet (DetQuant IndefArt number) cnf vpf = \e -> Exist (\x -> 
        And (And (cnf x) (Number number x)) (vpf x e)) ;
    iDet (DetQuant DefArt number) cnf vpf = \e -> Exist (\x -> 
        And (And (Unique (\z -> cnf x) x) (Number number x)) (vpf x e)) ;

fun iNP : NP -> (Ind -> Event -> Prop) -> Event -> Prop ;
def iNP (DetCN det cn) vp = iDet det (iCN cn) vp ;

fun iVP : VP -> Ind -> Event -> Prop ;
def
    iVP (UseV v)         = iV v ;
    iVP (ComplV2 v np) i = iNP np (\y -> iV2 v i y) ;
    iVP (AdvVP vp adv)   = iAdvVP adv (iVP vp) ;

fun iAdvVP : Adv -> (Ind -> Event -> Prop) -> Ind -> Event -> Prop ;
def iAdvVP (PrepNP by_Prep np) vpf = \i ->
        iNP np (\y,e -> And (vpf i e) (Agent y e)) ;

fun
    iV  : V  -> Ind        -> Event -> Prop ;
    iV2 : V2 -> Ind -> Ind -> Event -> Prop ;
def
    iV  v subj     e = And (VEvent v e) (Agent i e) ;
    iV2 v subj obj e = And (V2Event v e)
                           (And (Agent subj e) (Theme obj e)) ;
\end{verbatim} \end{small}
\end{small}

With these functions, a natural language abstract syntax tree of the sentence `a boy painted the houses':
\begin{small}
\begin{small} \begin{verbatim}
UseCl (TTAnt TPast ASimul) PPos (PredVP
    (DetCN (DetQuant IndefArt NumSg) (UseN boy_N))
    (ComplV2
        paint_V2
        (DetCN (DetQuant DefArt NumPl) (UseN house_N))))
\end{verbatim} \end{small}
\end{small}
can be interpreted as a logical semantics abstract syntax tree

\begin{small}
\begin{small} \begin{verbatim}
ExistEvent (\v0 -> And (And (Exist (\v1 -> And
        (And (iN boy_N v1) (Number NumSg v1))
        (Exist (\v2 -> And
            (And
                (Unique (\v3 -> iN house_N v2) v2)
                (Number NumPl v2))
            (And
                (V2Event paint_V2 v0)
                (And (Agent v1 v0) (Theme v2 v0)))))))
    (Anteriority ASimul v0)) (Time TPast v0))
\end{verbatim} \end{small}
\end{small}
which can then be linearised (see GitHub repository for the linearisation rules) into a simplified, flat format that resembles the COGS format:

\begin{small}
\begin{small} \begin{verbatim}
*house(v2); boy(v1) AND Number(v1, Sg) AND Number(v2, Pl) 
    AND paint(v0) AND Agent(v0, v1) AND Theme(v0, v2) 
    AND Anteriority(v0, Simul) AND Time(v0, Past)
\end{verbatim} \end{small}
\end{small}
In COGS, the semantics would have the following format:
\begin{small}
\begin{small} \begin{verbatim}
*house(x_4); boy(x_1) AND agent.paint(x_2, x_1) AND theme.paint(x_2, x_4)
\end{verbatim} \end{small}
\end{small}

\clearpage
\section{Distribution-based compositionality assessment of the SCAN variants} \label{sec:divergences}

\begin{figure*}[htb]
    \includegraphics[width=1\columnwidth]{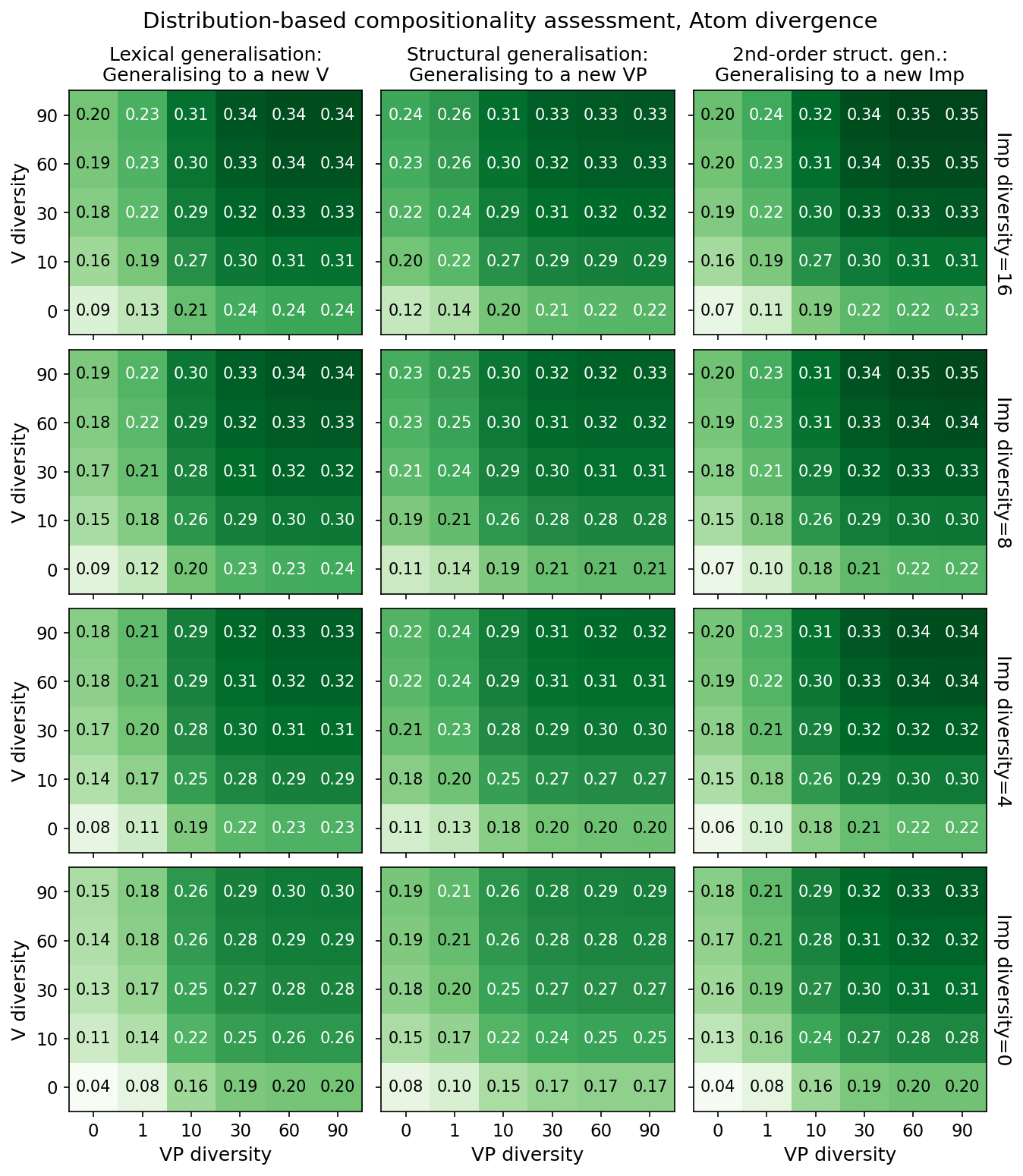}
    \caption{Atom divergences of the SCAN variants.}
    \label{fig:atom-div}
\end{figure*}

\begin{figure*}[htb]
    \includegraphics[width=1\columnwidth]{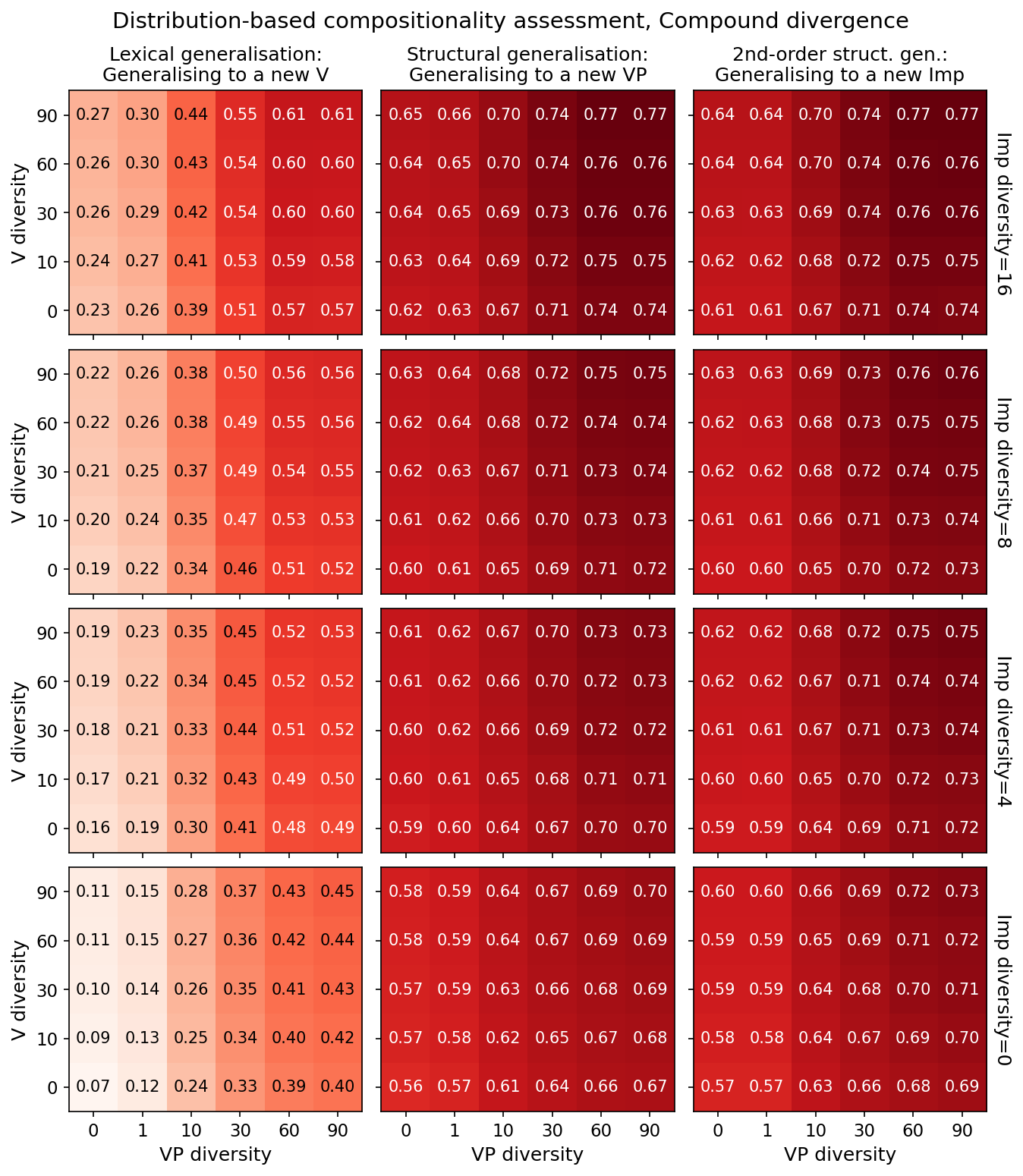}
    \caption{Compound divergences  of the SCAN variants.}
    \label{fig:compound-div}
\end{figure*}

Figures~\ref{fig:atom-div} and \ref{fig:compound-div} show the atom and compound divergences for all SCAN variant datasets. Divergence always correlates with type diversity.

\clearpage
\section{Ambiguity in the SLOG dataset} \label{sec:ambigous-slog}

In SLOG, the COGS fragment of English is extended with questions and relative clauses. These constructions create gaps, for example in the question `Whom did Noah give the cake to \_?', where `\_' denotes the place of the object noun phrase in the normal SVO sentence, e.g. `Noah gave the cake to Emily.', or in the relative clause: `Noah saw the cat that Emma gave a cake to \_.'

When a gapped construction is combined with the double object construction (DOC) of a ditransitive verb phrase,
the result is ambiguous since the gap may be in two different positions: `What did Emma feed the lion?' can be parsed as either `What did Emma feed the lion \_?', meaning the lion is the one doing the eating, or `What did Emma feed \_ the lion?', meaning the lion is being fed to something else. The latter parsing is used so rarely that it often doesn't sound grammatical, if the semantics don't support the parsing. When there is a strong semantic support for the gap-in-the-middle parsing, however, it sounds grammatical: `Whom did Emma tell \_ the story?'

The original COGS fragment includes potential ambiguity too, for example in the nested prepositional phrases: `Emma saw a cat in a house beside a table.' can be parsed so that the cat is beside a table or so that the house is beside a table; or one or both of the PPs could attach to the predicate. But these are systematically parsed so that the preposition attaches always to the immediately previous noun. 

However, in SLOG, the ambiguities created by the DOC and the gaps in questions and relative clauses are not so systematically resolved. The DOC with a gap is systematically used only with the gap after the last word, in the place of the direct object: `What did Emma feed the lion \_?' However, when there is a ditransitive verb in the relative clause as well as the main clause, and only one of them is DOC, we have two interpretations even if DOC has the gap always on the place of the direct object. For example,
`Whom did a girl give the cake that the boy forwarded a guest to?'
(a test sample in the `Wh-questions with modified NP' category) can be parsed in two ways:
\begin{quote}
Whom did a girl give [the cake that the boy forwarded a guest \_] to \_? \\
Whom did a girl give [the cake that the boy forwarded a guest to \_] \_?
\end{quote}
The first one is the semantically sensible interpretation, and in the second
the cake is the recipient of the giving and the forwarding. This is semantically unlikely, but resolving ambiguities based on semantic cues is not meant to be one of the challenges in SLOG, as there are very limited semantic cues for Transformers to exploit. Similarly `Emma sold a sandwich that the boy gave Emily to the girl.' (a training sample) can be parsed in two ways:
\begin{quote}
Emma sold [a sandwich that the boy gave Emily \_] to the girl. \\
Emma sold [a sandwich that the boy gave Emily to \_] the girl.
\end{quote}
although the latter is an unlikely scenario.

\section{Additional error analysis of SLOG experiments} \label{sec:discuss-ambiguity-variation}

We didn't include the Active subject questions of SLOG in the results because Transformers always achieve close to 100\% accuracy on this test sample group. However, there is a remaining few percent of the samples that Transformers systematically fail to interpret correctly.
The remaining difficulty comes from the unergative-unaccusative distinction of 1-place verbs (i.e. causative-inchoative alternation) combined with object alternation, i.e. verbs that may be 1- or 2-place verbs, such as eat and freeze. Subjects of unergative verbs are interpreted to be the agent of the event (`a cat eats'\ \ $\rightsquigarrow$ \ \ \texttt{cat(v1) AND eat(v0) AND Agent(v0, v1)}) whereas subjects of unaccusatives are the theme (`a cat freezes'\ \ $\rightsquigarrow$ \ \ \texttt{cat(v1) AND freeze(v0) AND Theme(v0, v1)}). In training, the verbs appear mostly as 2-place verbs, in which the subject is always the agent and never the theme (`a cat freezes the atmosphere'\ \ $\rightsquigarrow$ \ \ \texttt{atmosphere(v2); cat(v1) AND freeze(v0) AND Agent(v0, v1) AND Theme(v0, v2)}). The few remaining test samples that are systematically predicted incorrectly are of this type that include an unaccusative (not unergative) verb that appears mostly as a 2-place verb in the training set: 
`Who froze?'
`Who floated?'
`What rolled?'
`What decomposed?' etc.


\bibliographystyle{apalike}
\bibliography{custom,refs}

\end{document}